\documentclass[final,leqno,onefignum,onetabnum]{siamltex}

\pdfoutput=1
\usepackage{epsfig}
\usepackage{cite}
\usepackage{graphicx}
\usepackage{amsmath}
\usepackage{amssymb}
\usepackage{url}
\usepackage{paralist}
\usepackage[tight,footnotesize]{subfigure}
\usepackage{setspace}
\usepackage{mathdots}
\usepackage{cite}

\graphicspath{{img/}}

\newcommand{\bPhi}{{\bf \Phi}}
\newcommand{\bPsi}{{\bf \Psi}}

\newcommand{\bfH}{{\bf H}}
\newcommand{\bfU}{{\bf U}}
\newcommand{\bfD}{{\bf D}}
\newcommand{\bfI}{{\bf I}}
\newcommand{\bfZero}{{\bf 0}}
\newcommand{\bfF}{{\bf F}}

\newcommand{\bfy}{{\bf y}}
\newcommand{\bfx}{{\bf x}}
\newcommand{\bfz}{{\bf z}}
\newcommand{\bfs}{{\bf s}}
\newcommand{\bfe}{{\bf e}}

\newcommand{\bff}{{\bf f}}
\newcommand{\reals}{{\mathbb R}}

\newcommand{\static}{{\bf b}}

\newcommand{\secref}[1]{Sec.~\ref{#1}}
\newcommand{\figref}[1]{Fig.~\ref{#1}}

\providecommand{\inner}[2]{\ensuremath{\left\langle#1,#2\right\rangle}}

\title{Video Compressive Sensing for Spatial Multiplexing Cameras\\using Motion-Flow Models} 

\author{Aswin C. Sankaranarayanan\footnote{A. C. Sankaranarayanan is with the Department of Electrical and Computer Engineering at Carnegie Mellon University, Pittsbugh, PA. Email: saswin@ece.cmu.edu}, Lina Xu\footnote{L. Xu, Y. Li, K. F. Kelly and R. G. Baraniuk are with the Department of Electrical and Computer Engineering at Rice University, Houston, TX. Email: \{lx2, yun.li, kkelly, richb\}@rice.edu} , Christoph Studer\footnote{C. Studer is with the Department of Electrical and Computer Engineering at Cornell University, Ithaca, NY. Email: studer@cornell.edu}, Yun Li$^\dagger$, Kevin F. Kelly$^\dagger$, and Richard G. Baraniuk$^\dagger$}

\begin{document}
\maketitle
\newcommand{\slugmaster}{%
\slugger{siims}{2015}{xx}{x}{x--x}}

\begin{abstract}
Spatial multiplexing cameras (SMCs) acquire a (typically static) scene through a series of coded projections using a spatial light modulator (e.g., a digital micro-mirror device) and a few optical sensors. This approach finds use in imaging applications where full-frame sensors are either too expensive (e.g., for short-wave infrared wavelengths) or unavailable. 
Existing SMC systems reconstruct static scenes using techniques from compressive sensing (CS). For videos, however, existing acquisition and recovery methods deliver poor quality.
In this paper, we propose the CS multi-scale video (CS-MUVI)
sensing and recovery framework for high-quality video acquisition and recovery using SMCs. 
Our framework features novel sensing matrices that enable the efficient computation of a low-resolution video preview, while enabling high-resolution video recovery using convex optimization. 
To further improve the quality of the reconstructed videos, we extract optical-flow estimates from the low-resolution previews and impose them as constraints in the recovery procedure. 
We demonstrate the efficacy of our CS-MUVI framework for a host of synthetic and real measured SMC video data, and we show that high-quality videos can be recovered at roughly $60\times$ compression. 
\end{abstract}

\begin{keywords}
Video compressive sensing, optical flow, measurement matrix design, spatial multiplexing cameras
\end{keywords}

\begin{AMS}68U10, 68T45
\end{AMS}

\pagestyle{myheadings}
\thispagestyle{plain}
\markboth{Sankaranarayanan et al.}{CS-MUVI}

\section{Introduction} \label{sec:intro}
Compressive sensing (CS) enables one to sample signals  that admit a sparse representation in some transform basis well-below the Nyquist rate, while still enabling their faithful recovery \cite{candes2006robust, donoho2006compressed}. Since many natural and man-made signals exhibit a sparse representations, CS has the potential to reduce the costs associated with sampling in numerous practical applications.

\subsection{Spatial-multiplexing cameras}
The single pixel camera (SPC) \cite{duarte2008single} and its multi-pixel extensions \cite{wang2015lisens,chen2015fpa,Mahalanobis:14} are spatial-multiplexing camera (SMC) architectures that rely on CS.
In this paper, we focus on such SMC designs, which acquire random (or coded) projections of a (typically static) scene using a spatial light modulator (SLM) 
in combination with a small number of optical sensors, such as single photodetectors or bolometers.
The use of a small number of optical sensors---in contrast to full-frame sensors having millions of pixel elements---turns out to be advantageous when acquiring scenes at non-visible wavelengths. Since the acquisition of scene information beyond the visual spectrum often requires sensors built from exotic materials, corresponding full-frame sensor devices are either too expensive or cumbersome \cite{Gehm:15}.

\begin{figure}
\includegraphics[width=\textwidth]{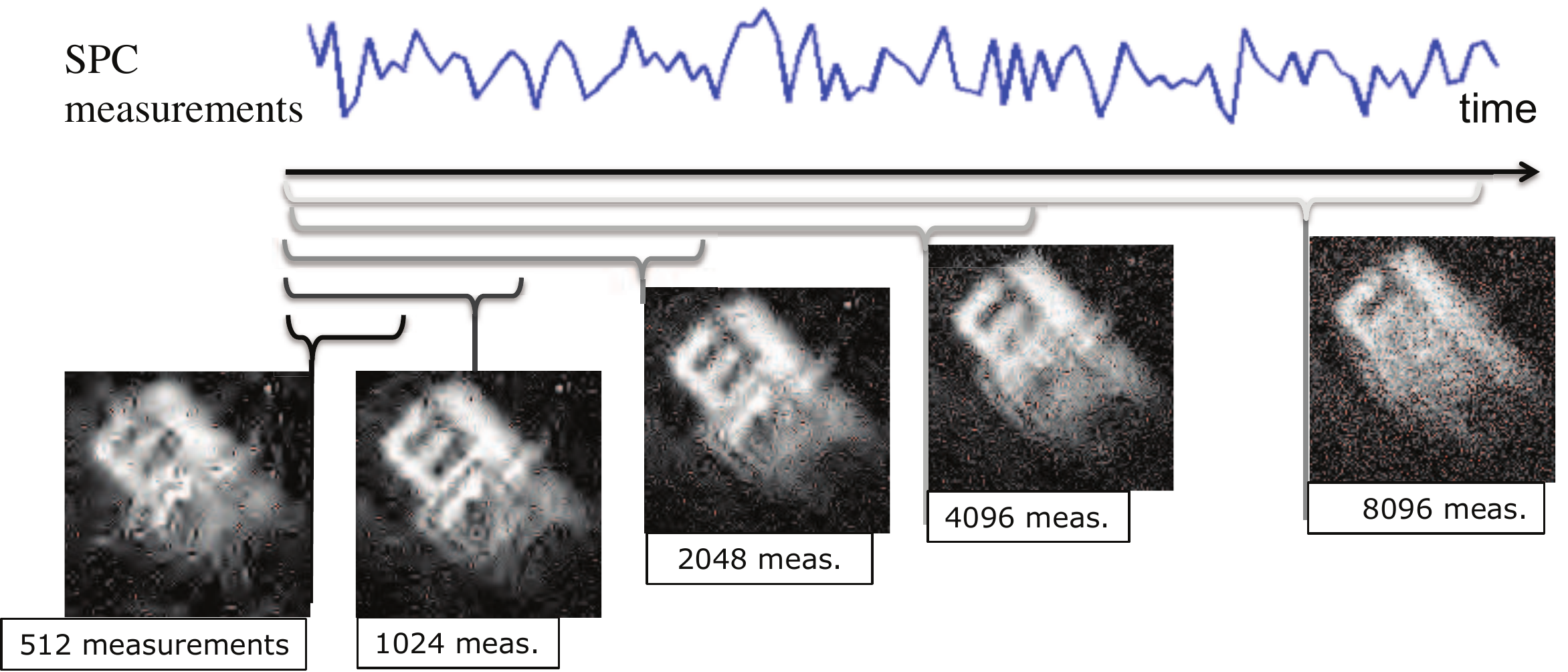}
\caption{\textbf{Single pixel camera (SPC) and the static scene assumption.} An SPC acquires a single measurement per time-instant. If the scene were static, one can aggregate multiple measurements over time to recover the image of the scene via sparse signal recovery; for dynamic scenes, however, this approach fails. 
Shown above are reconstructs of a scene comprising of a pendulum with the letter `R', swinging from right to left.
We show reconstructed images using different numbers of aggregated (or grouped) measurements. Aggregating only a small number of measurements, results in poor image quality. Aggregating a large number of measurements violates the static scene assumption and results in dramatic temporal aliasing artifacts.
}
\label{fig:static}
\end{figure}

Obviously, the use of a small number of sensors is, in general, not sufficient for acquiring complex scenes at high resolution. Hence, existing SMCs assume that the scenes to be acquired are static and acquire multiple measurements over time.
For static scenes (i.e., images) and for a single-pixel SMC architecture, this sensing strategy has been shown to deliver good results \cite{duarte2008single} typically at a compression of 2-8$\times$.
This approach, however, fails for time-variant scenes (i.e., videos). 
The main reason is due to the fact that the time-varying scene to be captured is ephemeral, i.e., \emph{each}  measurement acquires information of a (slightly) \emph{different} scene. The situation is further aggravated when we deal with SMCs having a very small number of sensors (e.g., only one for the SPC).
Virtually all existing methods for CS-based video recovery (e.g., \cite{park2009multiscale, sankaranarayanan2010compressive, vaswani2008kalman, wakin2006compressive, mun2011residual}) seem to overlook the important fact that scenes are changing while one acquires compressive measurements. 
In fact, all of the mentioned SMC video systems treat scenes as a sequence of \emph{static} frames (i.e., as piece-wise constant scenes) as opposed to a continuously changing scene.
This disconnect between the real-world operation of SMCs and the assumptions commonly made for video CS motivates novel SMC acquisition systems and recovery algorithms that are able to deal with the ephemeral nature of real scenes.
Figure~\ref{fig:static} illustrates the effect of assuming piece-wise static scenes. Put simply, grouping too few measurements for reconstruction results in poor spatial resolution; grouping too many measurements results in severe temporal aliasing artifacts. 


\subsection{The ``chicken-and-egg'' problem of video CS}
High-quality video CS recovery methods for camera designs relying on temporal multiplexing (in contrast to spatial multiplexing as it is the case for SMCs) are generally inspired by video compression schemes and exploit motion estimation between individually recovered frames \cite{reddy2011p2c2}.
Applying such techniques for SMC architectures, however, results in a fundamental problem:
On the one hand, obtaining motion estimates (e.g., the optical flow between pairs of frames) requires knowledge of the individual video frames. 
On the other hand, recovering the video frames in absence of motion estimates is difficult, especially when using low sampling rates and a small number of sensor elements (cf.~\figref{fig:static}).
Attempts to address this ``chicken-and-egg'' problem either perform multi-scale sensing  \cite{park2009multiscale} or sense separate patches of the individual video frames \cite{mun2011residual}. 
However, both approaches ignore the time-varying nature of real-world scenes and rely on a piecewise static scene model.

\begin{figure}[!ttt]
\includegraphics[width=\textwidth]{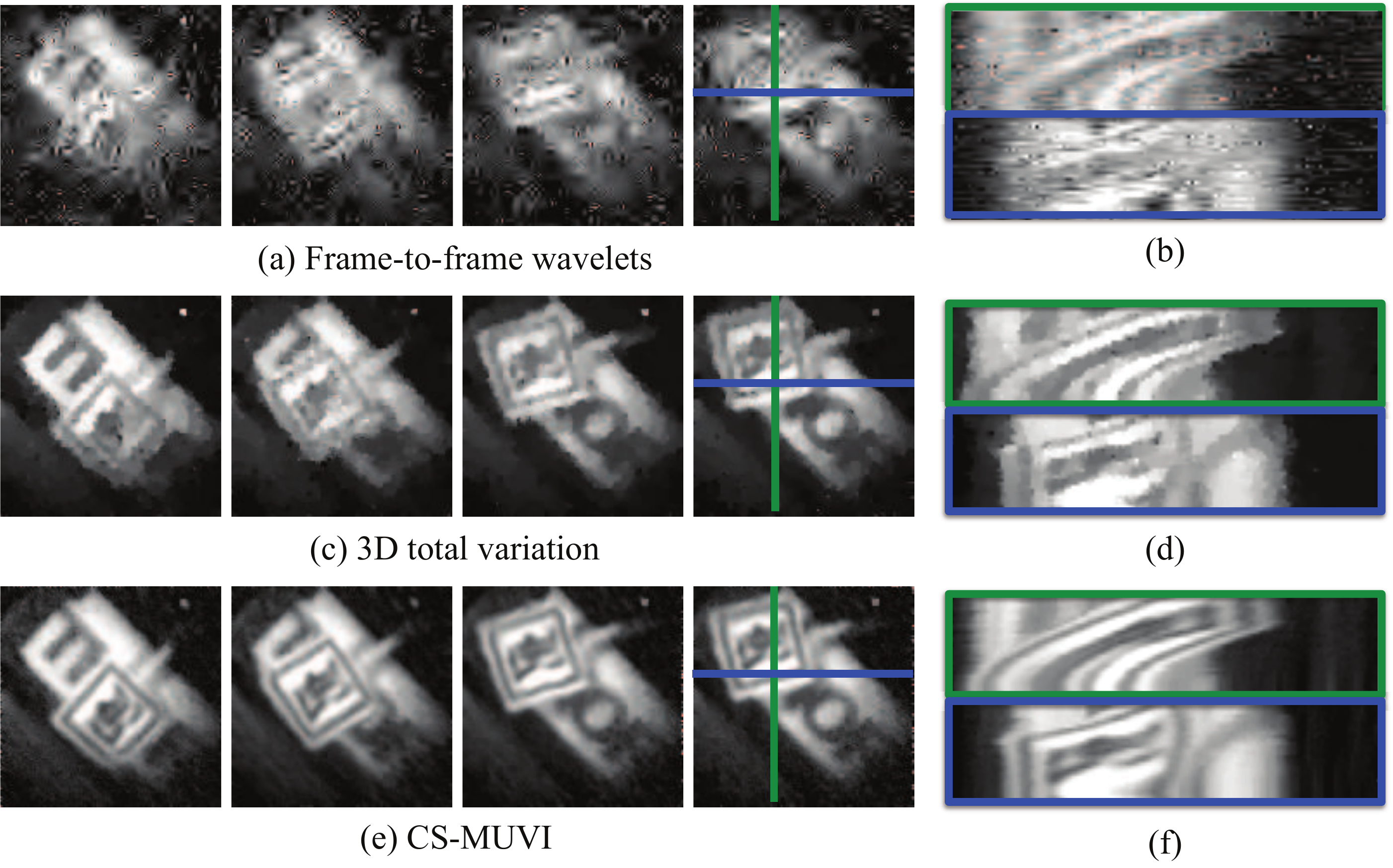}
\caption{\textbf{What a difference a signal model makes. }
We show  videos recovered from the same set of measurements but using different signal models: (a) sparsity of  wavelet coefficients of individual frames of the video, (b) 3D total variation enforcing sparse spatio-temporal gradients, and (c) CS-MUVI, the proposed video CS algorithm.
The data collected using an SPC operating in the short wave IR (SWIR) spectrum and acquiring  10,000 measurements/second at a spatial resolution of $128\times 128$ pixels. 
The scene, similar to Figure \ref{fig:static}, consists of a pendulum with the letter `R' swinging from right to left.
A  total of 16,384 measurements were acquired and videos were reconstructed under the three different signal models. Also shown are $xt$ and $yt$ slices corresponding to the lines marked. In all, CS-MUVI delivers high spatial as well as temporal resolution unachievable by both naive frame-to-frame wavelet sparsity as well as the more sophisticated 3D total variations model.  To the best of our knowledge, CS-MUVI is the first demonstration of successful video recovery at 128$\times$ super-resolution on {\em real data} obtained from an SPC. }
\label{fig:signalmodels}
\end{figure}

\subsection{The CS-MUVI framework}
In this paper, we propose a novel sensing and recovery method for videos acquired by SMC architectures, such as the SPC~\cite{duarte2008single}.
We start (in \secref{sec:overview}) with an overview of our sensing and recovery framework.
In \secref{sec:tradeoff},  we study the recovery performance of time-varying scenes and demonstrate that the performance degradation caused by violating the static-scene assumption is severe, even at moderate levels of motion. 
We then detail a novel video CS strategy for SMC architectures that overcomes the static-scene assumption. 
Our approach builds upon a co-design of scene acquisition and video recovery.  
In particular, we propose a novel class of CS matrices that enables us to obtain a low-resolution ``preview'' of the scene at low computational complexity. This preview video is used to extract robust motion estimates (i.e., the optical flow) of the scene at full-resolution (in \secref{sec:designMeas}).
We exploit these motion estimates to recover the full-resolution video by using off-the-shelf convex-optimization algorithms typically used for CS (in \secref{sec:optical}).
We demonstrate the performance and capabilities of our SMC video-recovery algorithm for a different scenes in \secref{sec:experiments}, show video recovery on real data in \secref{sec:real}, and discuss our findings in \secref{sec:discuss}.
Given the multi-scale nature of our framework, we refer to it as CS multi-scale video, or short CS-MUVI.
We note that a short version of this paper appeared at the IEEE International Conference on Computational Photography~\cite{sankaranarayanan2012cs} and Computational Optical Sensing and Imaging~\cite{xu2013multi} meeting. This paper contains an improved recovery algorithm, a more detailed performance analysis, and a larger number of experimental results. 
Most importantly, we show---to the best of our knowledge---the first high-quality video recovery results from real data obtained with a laboratory SPC; see \figref{fig:signalmodels} for corresponding results. 

\section{Background} \label{sec:prior}

\subsection{Design of multiplexing systems} \label{sec:hadamard}
Suppose that we have a signal acquisition system characterized by $\bfy = \mathbf{A} \bfx^* + \bfe,$ where $\bfx^* \in \reals^N$ is the signal to be sensed and $\bfy \in \reals^N$ is the measurement obtained using the  matrix $\mathbf{A}\in\reals^{N\times N}$. 
The entries $a_{ij}$ of the measurement matrix $\mathbf{A}\in\reals^{N\times N}$ are usually restricted to $a_{ij}\in[-1,+1]$. 
%
Given an invertible  matrix $\mathbf{A}$, the  recovery error  associated with the least-squares estimate $\widehat{\bfx} = \mathbf{A}^{-1} \bfy = \bfx^* + \mathbf{A}^{-1} \bfe$ 
satisfies the following inequality:
\[ \textit{ERR}(\widehat{\bfx}) =  \| \widehat{\bfx} - \bfx^* \|_2  \le  \| \mathbf{A}^{-1} \|  \| \bfe \|_2 . \]
Traditional imaging systems mostly use the identity as the measurement matrix, i.e.,  $\mathbf{A}={\mathbf I}_N$; such measurements result in an error equal to $\| \bfe \|_2$.

A classical problem is the design of  matrix $\mathbf{A}$, which results in minimal recovery error.
As shown in \cite{harwit1979hadamard}, Hadamard matrices are optimal in guaranteeing the smallest possible error when the measurement noise $\bfe$ is signal independent.
Specifically, if an $N \times N$ Hadamard matrix were to exist, then the recovery error satisfies $\textit{ERR}(\widehat{\bfx}) \le \| \bfe \|_2/\sqrt{N}$, which is a dramatic reduction from $\textit{ERR}(\widehat{\bfx}) \le \| \bfe \|_2$ achieved by $\mathbf{A}={\mathbf I}_N$.

%
While Hadamard multiplexing provides immense benefits in the context of imaging, it still requires an invertible measurement matrix, i.e, the dimensionality of the measurement $\bfy$ needs to be the same (or greater) than that of the sensed signal $\bfx^*$.
For SMCs that aggregate measurements over a time period, this implies a long acquisition period as the dimensionality of the signal $N$ increases. This also leads to a poorer temporal resolution. 
All of these concerns can potentially be addressed if it were possible to reconstruct a signal  from far-fewer measurements than its dimensionality  or when $M < N$.
Such a sensing framework is popularly referred to as \textit{compressive sensing.}
We discuss this approach next. 

\subsection{Compressive sensing} 
\label{sec:cs}

CS deals with the estimation of a  vector {$\bfx^* \in \reals^N$} from $M < N$ non-adaptive linear measurements  \cite{candes2006robust, donoho2006compressed}
\begin{align} \label{eqn:inputoutputrelation}
\bfy = \bPhi \bfx^*+ {\bfe},
\end{align} 
where $\bPhi \in \mathbb{R}^{M \times N}$ is the sensing matrix and ${\bfe}$ represents measurement noise.
Estimating the signal $\bfx^*$ from the compressive measurements $\bfy$ is an ill-posed problem, in general, since the (noiseless) system of equations $\bfy = \bPhi \bfx^*$ is under-determined.
Early results in sparse polynomial interpolation \cite{Ben-Or:1988:DAS:62212.62241} showed that, in the noiseless setting, it is possible to recover a $K$-sparse vector from $M = 2K$ measurements; however, the use of algebraic methods involving polynomials of high-degrees made the solutions  fragile to perturbations.
A fundamental result from CS theory states that a robust estimate of the vector $\bfx^*$ can be obtained from 
\begin{align} \label{eqn:reccondition}
M \sim K\log(N/K)
\end{align}
measurements if
\begin{inparaenum}[(i)] 
\item the signal $\bfx^*$ admits a $K$-sparse representation \mbox{$\bfs^* = \bPsi^T \bfx^*$} in an orthonormal basis $\bPsi$ (i.e., $\bfs^*$ has no more than $K$ non-zero entries), and 
\item the \emph{effective} sensing matrix $\bPhi\bPsi$ satisfies the restricted isometry property (RIP) \cite{candes2008restricted}.
\end{inparaenum}
For example, if the entries of the sensing matrix $\bPhi$ are i.i.d.\ zero-mean Gaussian distributed, then $\bPhi\bPsi$ is known to satisfy the RIP with high probability.
Furthermore, any $K$-sparse signal $\bfx^*$ satisfying \eqref{eqn:reccondition} can be estimated stably from the noisy measurement $\bfy$ by solving the following convex-optimization problem \cite{candes2006robust}:
\begin{align*}
\text{(P1)}\quad \widehat{\bfx}  =  \arg\min_{\bfx\in\reals^N} \| \bPsi^T \bfx \|_1 \quad \text{subject to}\,\,\| \bfy - \bPhi \bfx \|_2 \le \epsilon.
\end{align*}
Here, $(\cdot)^T$ denotes matrix transposition and the parameter $\epsilon \ge \| \bfe \|_2,$ is a bound on the measurement noise. For $K$-sparse signals, it can be shown that recovery error is bounded from above by 
$\textit{ERR}(\widehat{\bfx}) \le C_0 \epsilon$, where $C_0$ is a constant.
Hence, in the noiseless setting (where $\epsilon=0$), the $K$-sparse signal $\bfx^*$ can be recovered perfectly, even by acquiring far-fewer measurements \eqref{eqn:reccondition} than the signal's dimensionality. 
%


\paragraph{Signals with sparse gradients} 
The results of compressive sensing have been extended to include a broad class of signals beyond that of sparse signals; an example of this are signals that exhibit sparse gradients. 
For such signals, one can solve problems of the form \cite{chambolle2004algorithm,osher2005iterative}
\[  \text{(TV)} \quad \widehat{\bfx} = \arg\min_\bfx\,\textrm{TV}(\bfx)\quad \text{subject to}\,\,\| \bfy - \bPhi \bfx \|_2 \le \epsilon, \]
where the gauge $\textrm{TV}(\bfx)$ promotes sparse gradients. In the context of images where~$\bfx$ denotes a 2D signal (i.e., an image), the operator $\textrm{TV}(\bfx)$  can be defined as
\begin{align*}
\textrm{TV}_{\textrm{iso}}(\bfx) = \sum_i \sqrt{ (D_x \bfx(i))^2 + (D_y \bfx(i))^2 },
\end{align*}
where $D_x \bfx$ and $D_y\bfx$ are the spatial gradients in x- and y-direction of the 2-dimensional image $\bfx$, respectively. This definition can easily be extended to extended to higher-dimensional signals, such as RGB color images or videos (where the 3$^\text{rd}$ dimension is time). We next look at the prior art devoted specifically to CS of videos.

\subsection{Video compressive sensing}
An important challenge in  CS of videos is that the temporal dimension is fundamentally different from spatial and spectral dimensions due to its  ephemeral nature. 
The causality of time prevents us from obtaining additional measurements of an event that has already occurred.
This is especially relevant for SMCs that aggregate measurements over a time period. 
Further, temporal statistics of a video are often different from the spatial statistics.
These unique characteristics have lead to a large body of work dedicated to  video CS, and can broadly be grouped into signal models and corresponding recovery algorithms, and novel compressive imaging architectures.

\subsubsection{Spatial multiplexing cameras}
SMCs are imaging architectures that build on the ideas of CS. In particular, they employ an SLM, e.g., a digital micro-mirror device (DMD) or liquid crystal on silicon (LCOS), to optically compute a series linear projections of the scene~$\bfx$; these linear projections determine the rows of the sensing matrix $\bPhi$.
%
%
Since SMCs are usually built with only a few sensor elements, they can operate at wavelengths where corresponding full-frame sensors are too expensive.
In the recovery stage, one estimates the image $\bfx$ from the compressive measurements collected in~$\bfy$, for example, by solving (P1) or variants thereof.

\begin{figure}[!ttt]
\includegraphics[width=\textwidth]{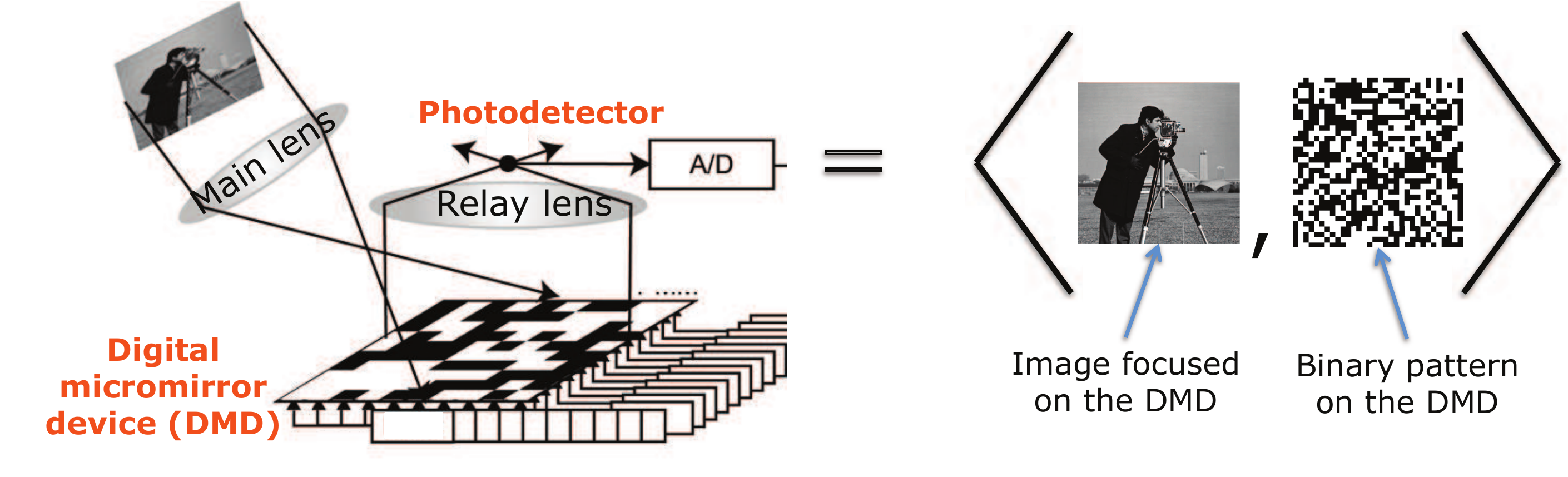}
\caption{\textbf{Operation principle of the single pixel camera (SPC).} Each measurement is the inner-product between the binary mirror-orientation patterns on the DMD and the scene to be acquired.}
\label{fig:spc}
\end{figure}

\paragraph{Single pixel camera} 
A prominent SMC is the SPC~\cite{duarte2008single};  
its main feature is the ability of acquiring images using only a \emph{single} sensor element (i.e., a single pixel) and by taking significantly fewer multiplexed measurements than the number of pixels of the scene to be recovered.
In the SPC, light from the scene is focused onto a programmable DMD, which directs  light from only a subset of activated micro-mirrors  onto the photodetector. The programmable nature of the DMD enables us to freely direct light from each of the micro-mirror towards the photodetector or away from it.
As a consequence, the voltage measured at the photodetector corresponds to an inner product of the image focused on the DMD and the activation pattern of the DMD (see Figure \ref{fig:spc}).
Specifically, at time $t$, if the DMD pattern were $\phi_t$ and the scene were $\bfx_t$, then the photodetector measures a scalar value $y_t = \langle \phi_t, \bfx_t \rangle + e_t$, where $\langle \cdot, \cdot \rangle$ denotes the inner-product between the vectors.
If the scene were static $\bfx_t = \bfx$, then multiple measurements can be aggregated to form the expression in~\eqref{eqn:inputoutputrelation}, with $\bPhi = [ \phi_1,\,\,\phi_2,\,\,\ldots\,\,\phi_M ]^T$.
The SPC leverages the high operating speed of the DMD, i.e., the mirror's orientation patterns on the DMD can be reprogrammed at kHz rates. The DMD's operating speed defines the measurement bandwidth (i.e., the number of measurements/second), which is one of the key factors that define the achievable spatial and  temporal resolutions. 

There have been many recovery algorithms proposed for video CS using the SPC. Wakin et al.\ \cite{wakin2006compressive} use 3-dimensional wavelets as a sparsifying basis for videos and recover all frames of the video jointly under this prior.
Unlike images, videos are not well represented using wavelets since they have additional temporal properties, like brightness constancy, that are better represented using motion-flow models.
Park and Wakin \cite{park2013multiscale} analyzed the coupling between spatial and temporal bandwidths of a video. In particular, they argue that reducing the spatial resolution of a scene implicitly reduces its temporal bandwidth and hence, lowers the error caused by the static scene assumption. 
This builds the foundation for the multi-scale sensing and recovery approach proposed in \cite{park2009multiscale},  where several compressive measurements are acquired at multiple scales for each video frame.
The recovered video at coarse scales (low spatial resolution) is used to estimate motion, which is  then used to boost the recovery at finer scales (high spatial resolution). 
%
%
Other scene models and recovery algorithms for video CS with the SPC use block-based models \cite{mun2011residual, fowler2010block}, sparse frame-to-frame residuals \cite{vaswani2010modified, cevher2008compressive}, linear dynamical systems \cite{vaswani2008kalman, sankaranarayanan2010compressive, sankaranarayanan2013compressive}, and low rank plus sparse models \cite{waters2011sparcs}.
%
%
To the best of our knowledge, all of them report results only on synthetic data and use the assumption that each frame of the video remains static for a certain duration of time (typically $1/30$ of a second)---an assumption that is violated when operating with an actual SPC.

\subsubsection{Temporal multiplexing cameras}
In contrast to SMCs that use sensors having low-spatial resolution and seek to spatially super-resolve images and videos, temporal multiplexing cameras (TMCs) have low frame-rate sensors and seek to temporally super-resolve videos.
In particular, TMCs use SLMs for temporal multiplexing of videos and sensors with high spatial resolution, such that the intensity observed at each pixel is coded temporally by the SLM during each exposure.

Veeraraghavan et al.~\cite{veeraraghavan2011coded} showed that periodic scenes could be imaged at very high temporal resolutions by using a global shutter or a ``flutter shutter'' \cite{raskar2006coded}. 
This idea was extended to non-periodic scenes in \cite{holloway2012flutter} where a union-of-subspace models was used to temporally super-resolve the captured scene.
Reddy et al.~\cite{reddy2011p2c2} proposed the per-pixel compressive camera (P2C2)  which extends the flutter shutter idea with per-pixel shuttering. 
Inspired from video compression standards such as MPEG-1 \cite{le1991mpeg} and H.264 \cite{richardson2004h}, the recovery of videos from the P2C2 camera was achieved using the optical flow between pairs of consecutive frames of the scene. 
The optical flow between pairs of video frames is estimated using an initial reconstruction of the high frame-rate video using wavelet priors on the individual frames.
A second reconstruction is then performed that further enforces the brightness constancy expressions provides by the optical flow fields.
The implementation of the recovery procedure described in \cite{reddy2011p2c2} is tightly coupled to the imaging architecture and prevents its use for SMC architectures.
Nevertheless, the use of optical-flow estimates for video CS recovery inspired the recovery stage of CS-MUVI as detailed in \secref{sec:optical}.

Gu et al.~\cite{gu2010coded} propose to use the rolling shutter of a CMOS sensor to enable higher temporal resolution. The key idea there is to stagger the exposures of each row randomly and use image/video statistics to recover a high-frame rate video.
Hitomi et al.~\cite{hitomi2011video} uses a per-pixel coding, similar to  P2C2, that is implementable in modern CMOS sensors with per-pixel electronic shutters; however, a hallmark of their approach is the use of a highly over-complete dictionary of video patches to recovery the video at high frame rates. This results in highly accurate reconstructions even when brightness constancy---the key construct underlying optical flow estimation---is violated.
Llull et al.\ \cite{llull2013coded} propose a TMC that uses a translating mask in the sensor plane to achieve temporal multiplexing. This approach avoids the hardware complexity involved with DMDs and LCOS, and enjoys other benefits including low operational power consumption. 
In Yang et al.\ \cite{yang2014video}, a Gaussian Mixture Model (GMM) is used as a signal prior to recovery high-frame rate videos for TMCs; a hallmark of this approach is that the GMM parameters are not just trained offline but also adapted and tuned in situ during the recovery process.
Harmany et al.\ \cite{harmany2013compressive} extend coded aperture systems by incorporating a flutter shutter \cite{raskar2006coded} or a coded exposure; the resuling TMC provides immense flexibility in the choice of measurement matrix.
They  also show the resulting system provides measurement matrices that satisfy the RIP. 


%

\section{Overview of CS-MUVI} \label{sec:overview}

State-of-the-art video compression methods rely on estimating the motion in the scene, compress a few reference frames, and use the motion vectors that relate the remaining parts of a scene to these reference frames. While this approach is possible in the context of video compression, i.e., where the algorithm has prior access to the entire video, it is significantly more difficult in the context of compressive sensing.

A general strategy to enable the use of motion flow-based signal models for video CS is to use a two-step  approach \cite{reddy2011p2c2}.
In the first step, an initial estimate of the video is generated by recovering each frame individually using sparse wavelet  or gradient priors.
The initial estimate is used to derive motion flow between consecutive frames; this enables a powerful description in terms of relating  intensities at pixels across frames.
In the second step, the video is re-estimated but now with the aid of enforcing the extracted motion flow constraints in addition to the measurement constraints.
%
The success of this two step strategy critically depends on the ability to obtain reliable motion estimates, which, in turn, depends on obtaining robust initial estimates in the first step.
Unfortunately, in the context of SMCs, obtaining reliable initial estimates of the frames of the video, in absence of motion knowledge, is inherently hard due to the violation of the static scene model (recall \figref{fig:static}).

The proposed framework, referred to as CS-MUVI, enables a robust initial estimate by obtaining the individual frames at a \textit{lower spatial resolution}.
This approach has two important benefits towards reducing the violation of the static scene model.
First, obtaining the initial estimate at a lower spatial resolution reduces the dimensionality of the video significantly. 
As a consequence, we can estimate individual frames of the video from \textit{fewer} measurements.
In the context of an SMC, this implies a \textit{smaller} time window over which these measurements are obtained, and hence, \textit{reduced} misfit to the static scene model.
Second, spatial downsampling naturally reduces the temporal resolution of the video \cite{park2013multiscale}; this is a consequence of the additional  blur due to spatial-downsampling.
This implies that the violation of the static scene assumption is naturally \textit{reduced} when the video is downsampled.
In \secref{sec:tradeoff}, we study this strategy in detail and characterize the error in estimating the initial estimates at a lower resolution.
Specifically, given $W$ consecutive measurements from an SMC, we are interested in estimating a \textit{single static} image at a resolution of $\sqrt{W} \times \sqrt{W}$ pixels.
Note that varying $W$, which denotes the window length, varies both the spatial resolution of the recovered frame (since it has a resolution of $\sqrt{W} \times \sqrt{W}$) as well as its temporal resolution (since the acquisition time is proportional to $W$).
We analyze various sources of error in the recovered low-resolution frame.
This analysis provides conditions for stable recovery of the initial estimates that leads to the design of measurement matrices in \secref{sec:designMeas}.

\begin{figure*}[t]
\centering
\includegraphics[width=\textwidth]{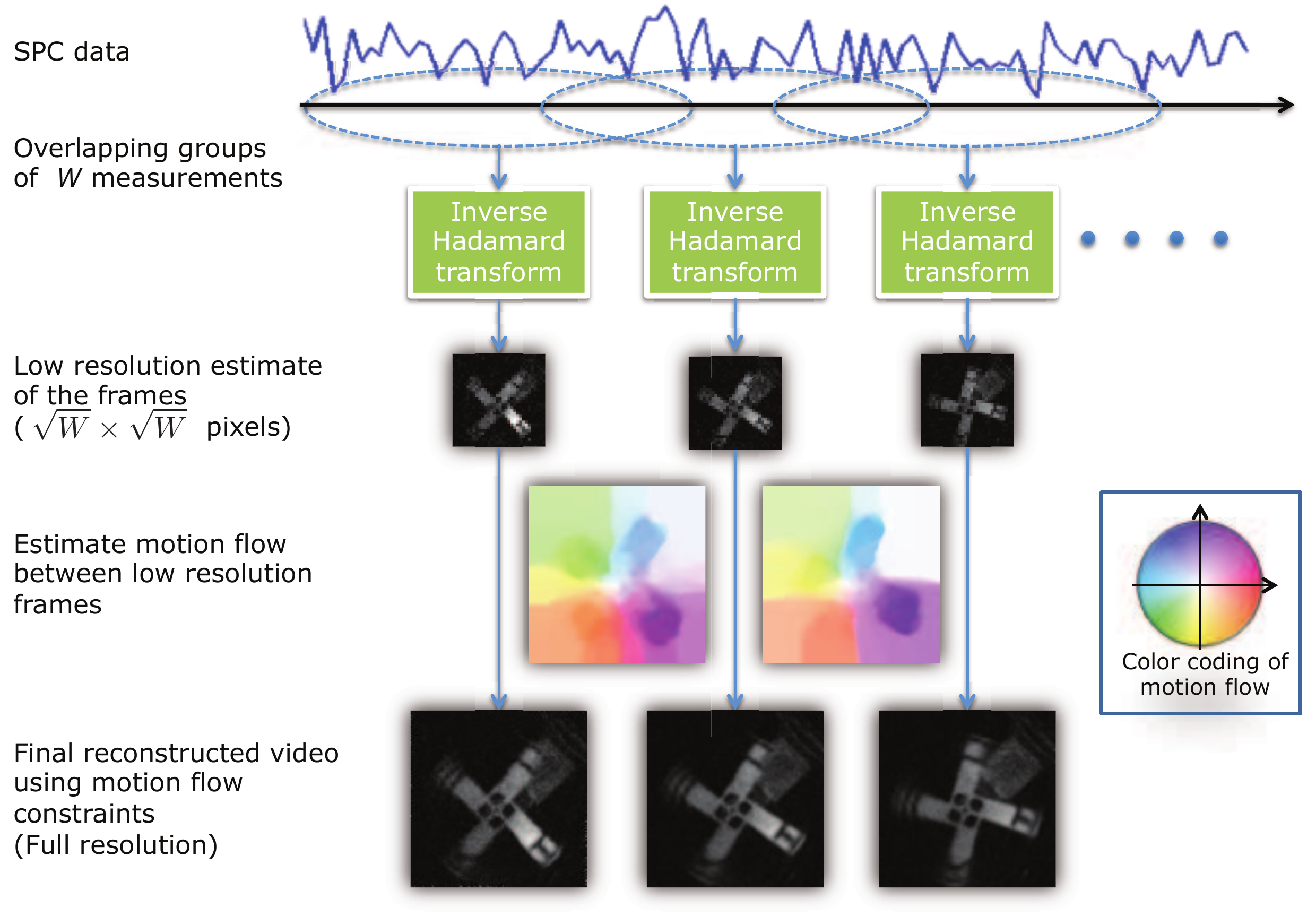}
\caption{\textbf{Outline of the CS-MUVI recovery framework.} Given a total number of $T$ measurements, we group them into overlapping windows of size $W$ resulting in a total of $F$ frames.
For each frame, we first compute a low-resolution initial estimate using a window of $W$ neighboring measurements. We then compute the optical flow between upsampled preview frames (the optical flow is color-coded as in \cite{liu2009beyond}).
Finally, we  recover $F$ high-resolution video frames by enforcing a sparse gradient prior along with the measurement constraints, as well as the brightness constancy constraints generated from  the optical-flow estimates.}
\label{fig:OpFlowDiag}
\end{figure*}

The proposed CS-MUVI framework for video CS relies on three steps.
First, we recover a low-resolution video by reconstruction each frame of the video, individually, using simple least-squares techniques.
Second, this low-resolution video is used to obtain motion  estimates between frames.
Third, we recover a high-resolution video by enforcing a spatio-temporal gradient prior, the constraints induced by the compressive measurements as well as the constraints due to motion estimates.
\figref{fig:OpFlowDiag} provides an overview schematic of these steps.

\section{Spatio-temporal trade-off} \label{sec:tradeoff}

We now study the recovery error that results from the static-scene assumption while sensing a time-varying scene (video) with an SMC. We also identify a fundamental trade-off underlying a multi-scale recovery procedure, which is used in \secref{sec:designMeas} to identify novel sensing matrices that minimize the spatio-temporal recovery errors.
Since the SPC is the most challenging SMC architecture as it only provides a single pixel sensor, we solely focus on the SPC in the following. Generalizing our results to other SMC architectures with more than one sensor is straightforward.

\subsection{SMC acquisition model}

The compressive measurements $y_t \in \reals$ taken by a single-pixel SMC at the sample instants~$t=1,\ldots,T$ can be modeled as 
\begin{align*}
y_t = \inner{\phi_t}{\bfx_t} + e_t, 
\end{align*}
where  $T$ is the total number of acquired samples, $\phi_t \in \reals^{N \times 1}$ is the measurement vector, $e_t \in \reals$ represents measurement noise, and $\bfx_t \in \reals^{N \times 1}$ is the scene (or frame) at sample instant~$t$.
In the remainder of the paper, we assume that the 2-dimensional scene consists of $n \times n$ spatial pixels, which, when vectorized, results in the vector $\bfx_t$ of dimension $N = n^2$.
We also use the notation $\bfy_{1:W}$ to represent the vector consisting of a window of $W\leq{}T$ successive compressive measurements (samples), i.e,
\begin{align} \label{eqn:measStackTime}
 \bfy_{1:W}  = \left[ \begin{array}{c} y_1 \\ y_{2} \\ \vdots \\y_W \end{array} \right]
= \left[ \begin{array}{c} 
\inner{\phi_1}{\bfx_1} + e_1 \\ 
\inner{\phi_2}{\bfx_2} + e_2 \\ 
\vdots \\ 
\inner{\phi_W}{\bfx_W} + e_W 
\end{array} \right].
\end{align}

\subsection{Static-scene and down-sampling errors}
\label{sec:errorsources}

Suppose that we rewrite our (time-varying) scene~$\bfx_t$ for~a window of~$W$ consecutive sample instants as follows:
\begin{align*}
\bfx_t = \static + \Delta \bfx_t,  \quad t=1,\ldots,W.
\end{align*} 
Here, $\static$ is the static component (assumed to be invariant for the considered window of $W$ samples), and $\Delta \bfx_t=\bfx_t-\static$ is the error at sample instant $t$ caused by the static-scene assumption. 
By defining \mbox{$z_t = \inner{\phi_t}{\Delta \bfx_t}$}, we can rewrite~\eqref{eqn:measStackTime} as
\begin{equation} \label{eqn:tempError}
\bfy_{1:W}  = \bPhi \static + \bfz_{1:W} +  \bfe_{1:W} ,
\end{equation}
where $\bPhi \in \reals^{W \times N}$ is the sensing matrix whose $t$-th row corresponds to the transposed measurement vector $\phi_t$.

We now investigate the error caused by spatial downsampling of the static component $\static$ in \eqref{eqn:tempError}.
To this end, let $\static_L \in \reals^{N_L}$ be the down-sampled static component, and assume $N_L = n_L \times n_L$ with $N_L<N$. 
By defining a linear up-sampling and down-sampling operator as $\bfU \in \reals^{N\times N_L}$ and $\bfD \in \reals^{N_L\times N}$, respectively, we can rewrite \eqref{eqn:tempError} as follows: 
 \begin{align} 
 \bfy_{1:W}  & = \bPhi  (\bfU \static_L +  \static - \bfU \static_L) + \bfz_{1:W}  + \bfe_{1:W} \notag \\
 & = \bPhi \bfU \static_L + \bPhi ( \static - \bfU \static_L )  + \bfz_{1:W}  + \bfe_{1:W} \notag \\
 & = \bPhi \bfU \static_L + \bPhi ( \bfI -\bfU\bfD ) \static + \bfz_{1:W}  + \bfe_{1:W} \label{eqn:spaceTime}
\end{align}
since $\static_L = \bfD \static$.
Inspection of \eqref{eqn:spaceTime} reveals three sources of error in the CS measurements of the low-resolution static scene $\bPhi \bfU \static_L$:
\begin{inparaenum}[(i)]
\item The \emph{spatial-approximation error}  \mbox{$\bPhi ( \bfI -\bfU\bfD ) \static$} caused by down-sampling,
\item the \emph{temporal-approximation error} $\bfz_{1:W}$ caused by assuming the scene remains static for $W$ samples, and  
\item the \emph{measurement error}~$\bfe_{1:W}.$
\end{inparaenum}
Note that when $W \ge N_L$, the matrix $\Phi U$ has at least as many rows as columns and hence, we can get an estimate of $\static_L = (\Phi U)^{\dagger} \bfy_{1:W}$.
We next study  the error induced by this least-squares estimate in terms of the relative contributions of the spatial-approximation and temporal-approximation terms.

\subsection{Estimating a low-resolution image}
\label{sec:lowresimage}

In order to analyze the trade-off that arises from the static-scene assumption and the down-sampling procedure,  we consider the scenario where the effective matrix $\bPhi \bfU$ is of dimension $W\times{}N_L$ with $W\geq N_L$; that is, we aggregate at least as many compressive samples as the down-sampled spatial resolution. 
If $\bPhi \bfU$ has full (column) rank, then we can obtain a least-squares (LS) estimate $\widehat{\static}_L$ of the low-resolution static scene $\static_L$ from \eqref{eqn:spaceTime} as
\begin{align} \label{eqn:spaceTimeinverted}
\widehat{\static}_L &= \left( \bPhi \bfU\right)^{\dagger} \!\bfy_{1:W}  = \static_L +  \left( \bPhi \bfU\right)^{\dagger} \!\big( \bPhi ( \bfI -\bfU\bfD ) \static + \bfe_{1:W}  + \bfz_{1:W} \big)
\end{align}
where $\left(\cdot\right)^{\dagger}$ denotes the pseudo inverse.
From~\eqref{eqn:spaceTimeinverted} we observe the following facts:
\begin{inparaenum}[(i)]
\item The window length~$W$ controls a trade-off between the spatial-approximation error \mbox{$\bPhi ( \bfI -\bfU\bfD ) \static$} and the error $\bfz_{1:W}$ induced by assuming a static scene $\static$, and 
\item the least squares (LS) estimator matrix $\left( \bPhi \bfU\right)^{\dagger}$ (potentially) amplifies all three error sources.
\end{inparaenum}

\subsection{Characterizing the trade-off}
\label{sec:spatiotemporal}

\begin{figure}[t]
\centering
\subfigure[Synthetic video of a translating object over a static textured background]{
\includegraphics[width=0.8\columnwidth]{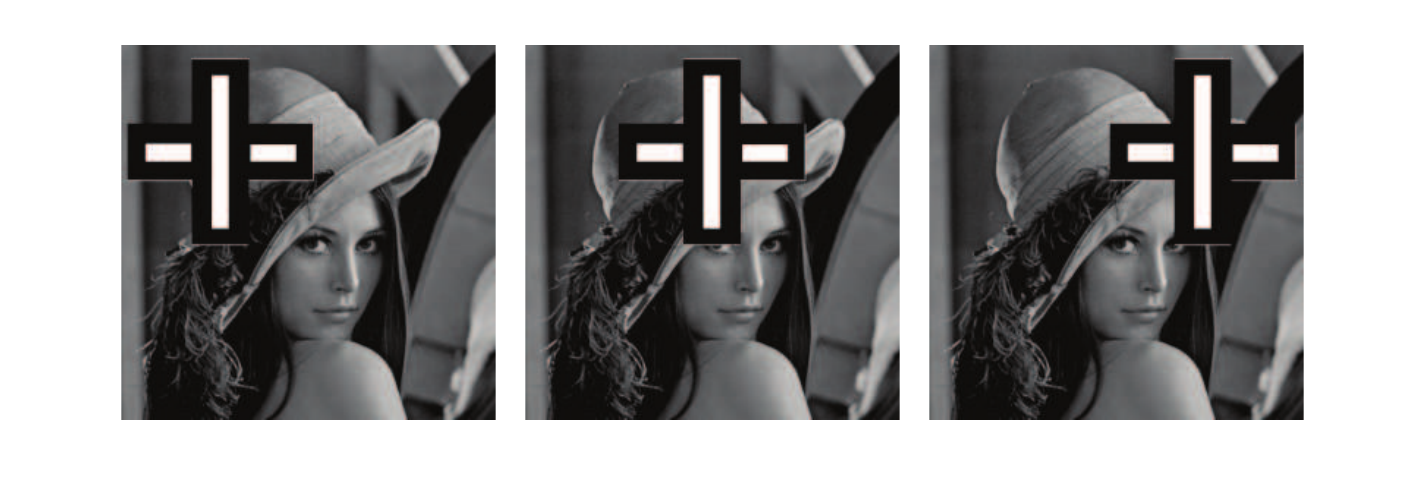}
\label{fig:lenacross}
}
\subfigure[separate error sources]{
\includegraphics[width=0.472\columnwidth]{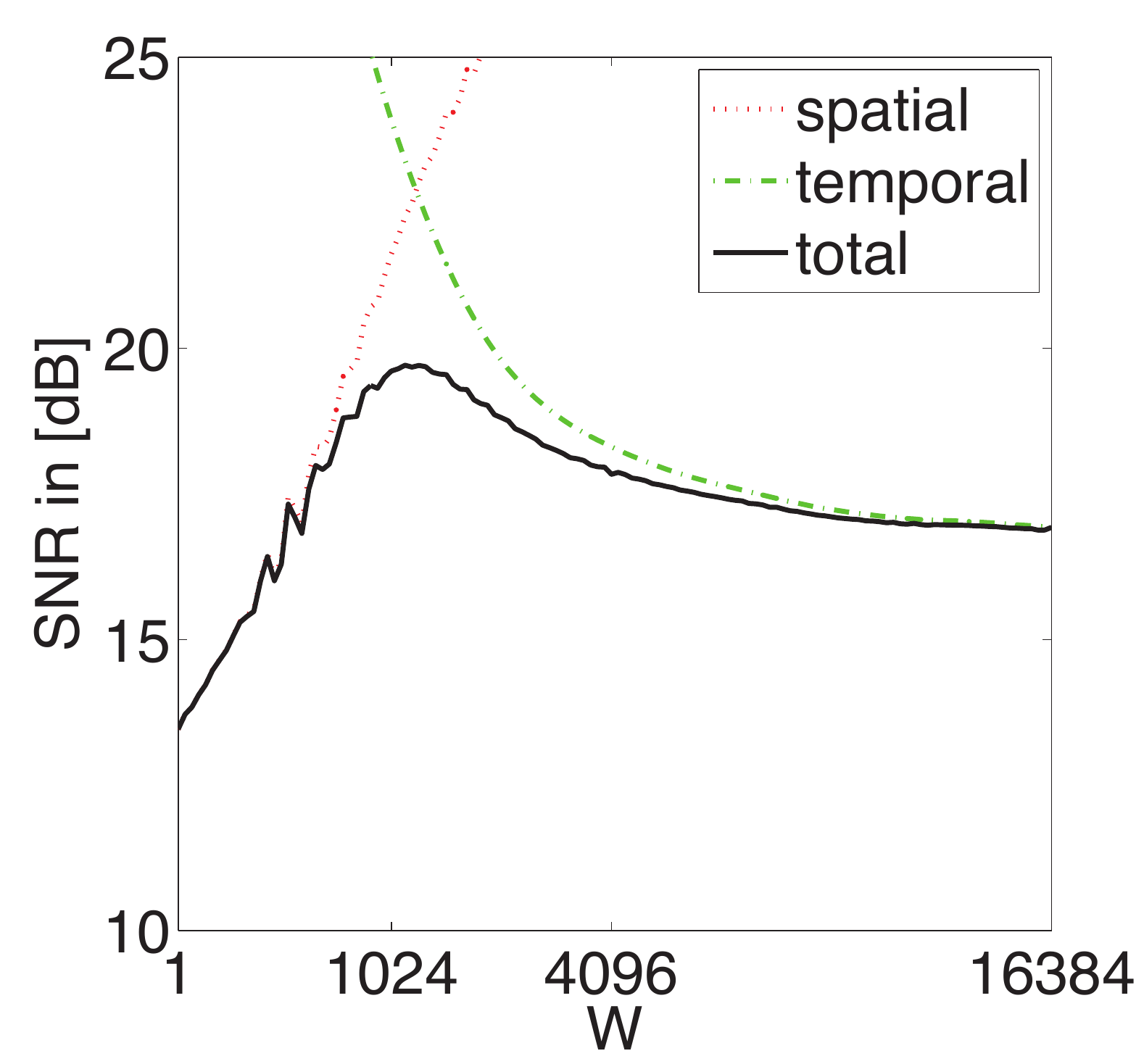}
\label{fig:spaceVtimea}
}
\subfigure[impact of temporal changes]{
\includegraphics[width=0.472\columnwidth]{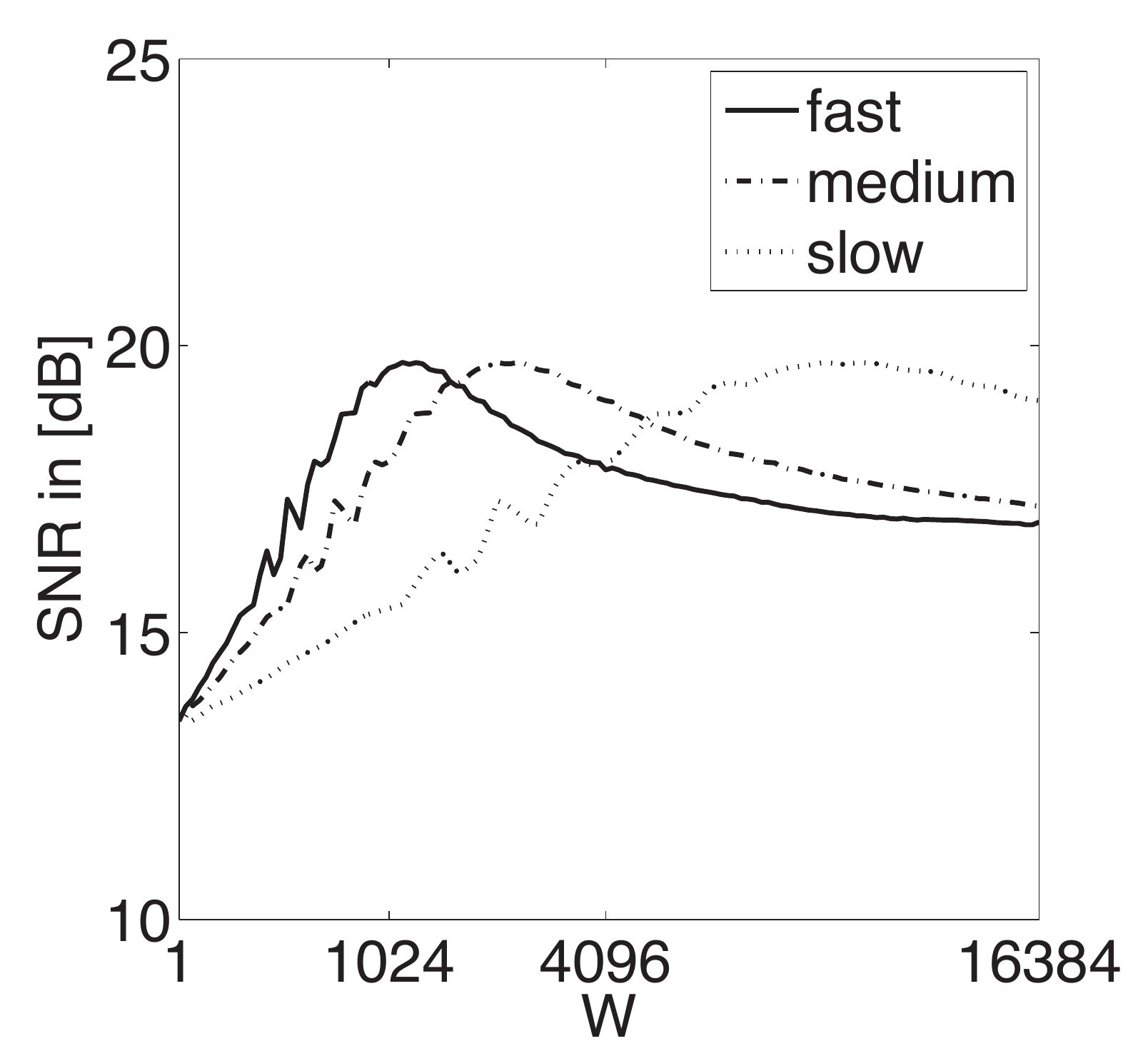}
\label{fig:spaceVtimeb}
}
\caption{\textbf{Trade-off between spatial and temporal approximation errors.} The plots corresponding to a scene with a translating object over a static background. (a) Frames of a synthetic video with a spatial resolution of $128\times 128$ pixels. The speed of movement of the cross is precisely controlled to sub-pixel accuracy.  (b) The recovery SNRs caused by spatial and temporal approximation errors for values of $W$, the total number of measurements obtained. We collect $W = n_L^2$ measurements under the measurement model in (\ref{eqn:measStackTime}), and reconstruct a single static frame $\widehat{\static}_L$ at a resolution of $n_L \times n_L$, such that $(\bPhi U)$ is invertible, using (\ref{eqn:spaceTimeinverted}). Next, since we have the ground truth, we can independently compute the spatial error $\| \static - \static_L \|$ as well as the temporal error $\| \bfz_{1:W} \|$.  (c) We can vary the speed of motion of object and observe the dependence of the total approximation error on the speed of the object. At the medium speed, the cross translates so as the cover the field-of-view within $16,384$ measurements; the speed of translation for the `slow' and `fast' motions correspond to one-half and twice the speed of translation at `normal', respectively.} 
\label{fig:spaceVtime}
\end{figure}

The spatial approximation error and the temporal approximation error are both  functions of the window length $W$. We now show that carefully selecting $W$ minimizes the combined spatial and temporal error in the low-resolution estimate $\widehat{\static}_L$.
A close inspection of \eqref{eqn:spaceTimeinverted} shows that for $W = 1$, the temporal-approximation error is zero, since the static component $\static$ is able to perfectly represent the scene at each sample instant $t$.
As $W$ increases, the temporal-approximation error increases for time-varying scenes; simultaneously, increasing $W$  reduces the error caused by down-sampling $\bPhi ( \bfI -\bfU\bfD ) \static$ (see \figref{fig:spaceVtimea}). For $W\geq N$  there is no spatial approximation error (as long as $\bPhi\bfU$ is invertible). 
Note that characterizing both errors analytically is, in general, difficult as they heavily depend on the  on the scene under consideration.

Figure \ref{fig:spaceVtime} illustrates the trade-off controlled by $W$ and the individual spatial and temporal approximation errors, characterized in terms of the recovery signal-to-noise-ratio (SNR).
The figure highlights our key observation that there is an optimal window length~$W$ for which the total recovery SNR is maximized. In particular, we see from \figref{fig:spaceVtimeb} that the optimum window length increases (i.e., towards higher spatial resolution) when the scene changes slowly; in contrary, when the scene changes rapidly, the window length (and consequently, the spatial resolution) should be low. 
Since $N_L\leq W$, the optimal window length~$W$ dictates the resolution for which accurate low-resolution motion estimates can be obtained.
Hence, the optimal window length depends on the scene to be acquired, the rate of which measurements can be acquired, and the sensing matrix $\bPhi$ itself.

\section{Design of sensing matrix} \label{sec:designMeas}
%

\sloppy
In order to bootstrap CS-MUVI, a low-resolution estimate of the scene is required. We next show that carefully designing the CS sensing matrix $\bPhi$ enables us to compute high-quality low-resolution scene estimates at low complexity, which improves the performance of video recovery.

\subsection{Dual-scale sensing matrices}

\begin{figure}[!ttt]
\centering
\includegraphics[width=\columnwidth]{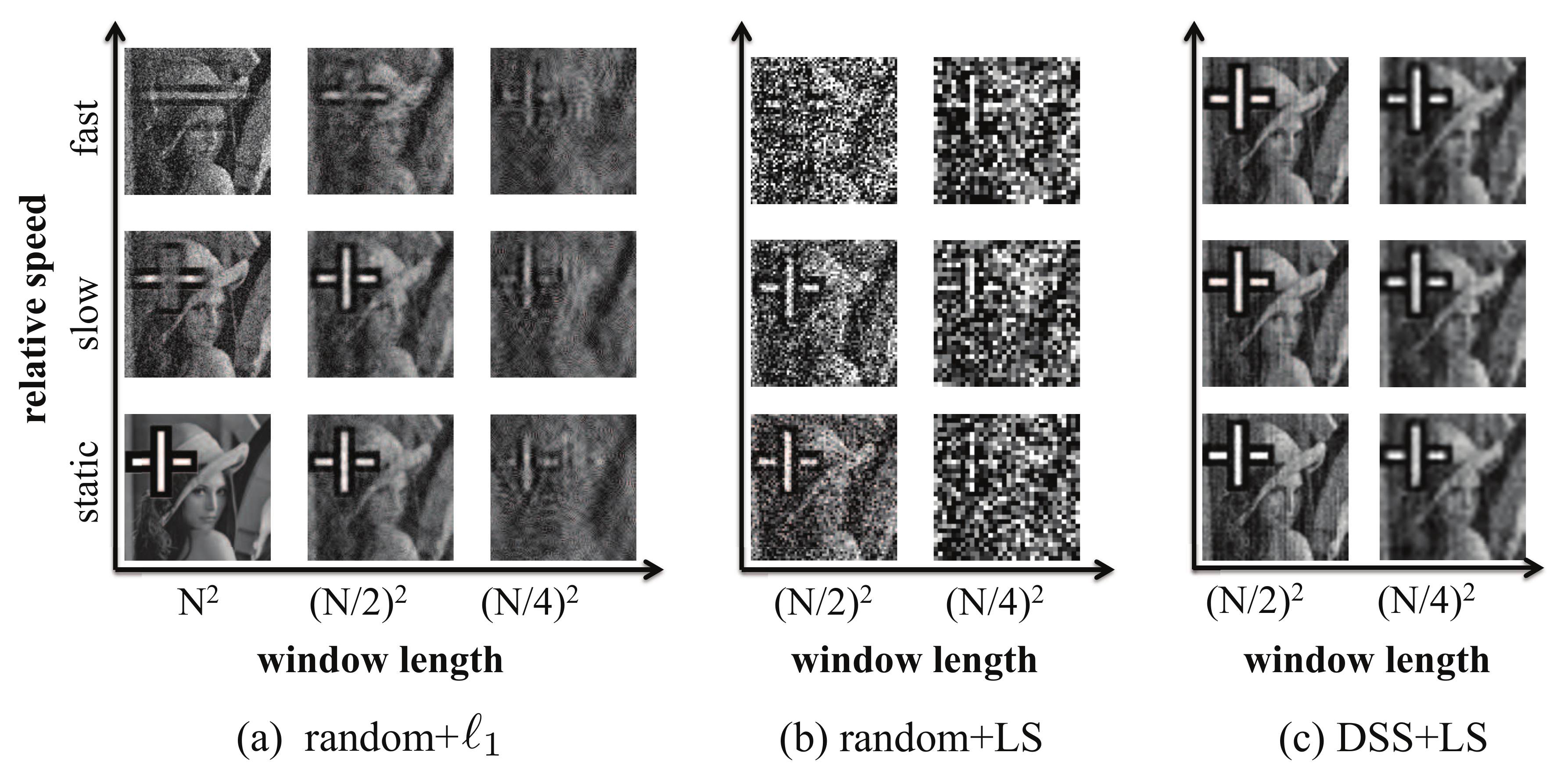}
\caption{\textbf{Performance of $\ell_1$ and $\ell_2$-based recovery algorithms for varying object motion.} The underlying scene corresponds to translating cross over a static background of Lena.
The speed of translation of the cross is varied across different rows. Comparison between (a) $\ell_1$-norm recovery, (b) LS recovery using a random matrix, and (c) LS recovery using a dual-scale sensing (DSS) matrix  for various relative speeds (of the cross) and window lengths $W$.}
\label{fig:LenaL1_CompMotion}
\end{figure}

The choice of the sensing matrix $\bPhi$ and the upsampling operator $\bfU$ are critical to arrive at a high-quality estimate of the low-resolution image $\static_L$.
Indeed, if the effective matrix $\bPhi \bfU$ is ill-conditioned, then application of the pseudo-inverse $(\bPhi \bfU)^\dagger$ amplifies all three sources of errors in~\eqref{eqn:spaceTimeinverted}, eventually resulting in a poor estimate. 
For virtually all sensing matrices $\bPhi$ commonly used in CS, such as i.i.d.\ (sub-)Gaussian matrices, as well as sub-sampled Fourier or Hadamard matrices, right multiplying them with an upsampling operator~$\bfU$ often results in an  ill-conditioned matrix or even a rank-deficient matrix. Hence, well-established CS matrices are a poor choice for obtaining a high-quality low-resolution preview. 
Figures \ref{fig:LenaL1_CompMotion}(a) and \ref{fig:LenaL1_CompMotion}(b) show recovery results for na\"ive recovery using (P1) and least-squares (LS), respectively, using a random  sensing matrix. We immediately see that both recovery methods result in poor performance, even for large window sizes~$W$ or for a small amount of motion.

In order to achieve good CS recovery performance \emph{and} have minimum noise enhancement when computing a low-resolution preview $\widehat\static_L$ according to \eqref{eqn:spaceTimeinverted}, we propose a novel class of sensing matrices, referred to as \emph{dual-scale sensing} (DSS) matrices. 
These matrices will (i) satisfy the RIP to enable CS and (ii) remain well-conditioned when right-multiplied by a given up-sampling operator $\bfU$.
Such a DSS matrix enables robust low-resolution as shown in  \figref{fig:LenaL1_CompMotion}(c).
We next discuss the details.
%
%
%
%

\subsection{DSS matrix design}
In this section, we detail a particular design that is suited for SMC architectures. 
In SMC architectures, we are constrained in the choice of the entries of the sensing matrix~$\bPhi$. Practically, the DMD limits us to matrices having binary-valued entries (e.g., $\pm 1$) if we are interested in the highest possible measurement rate.\footnote{It is possible to employ more general sensing matrices, e.g., using spatial and/or temporal half-toning, which, however, comes at the cost of spatial resolution and/or speed. The design of such matrices are not in the scope of this paper but an interesting research direction.}
We propose the matrix $\bPhi$ to satisfy $\bfH = \bPhi\bfU$, where $\bfH$ is a $W\times{}W$ Hadamard matrix\footnote{In what follows, we assume that $W$ is chosen such that a $W\times{}W$ Hadamard matrix exists.} and $\bfU$ is a predefined up-sampling operator.
Recall from Section \ref{sec:hadamard},  Hadamard matrices have the following advantages: 
\begin{inparaenum}[(i)]
\item they have orthogonal columns, 
\item they exhibit optimal SNR properties over matrices restricted to $\{-1,+1\}$ entries, and
\item applying the (forward and inverse) Hadamard transform requires very low computational complexity (i.e., the same complexity as a fast Fourier transform).
\end{inparaenum}

\begin{figure}
\center
\includegraphics[width=\columnwidth]{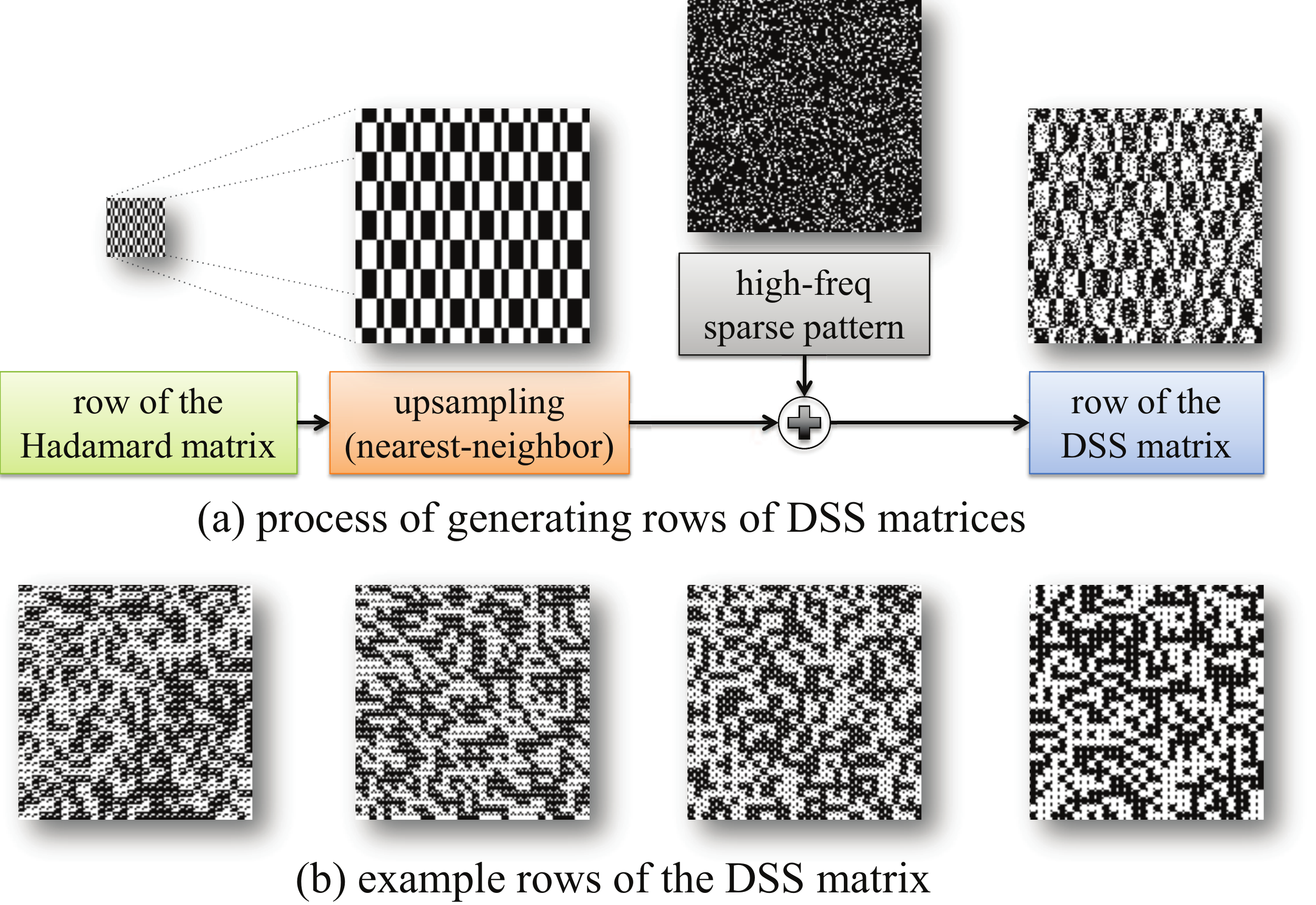}
\caption{{\bf Generating DSS patterns.} (a) Outline of the process in \eqref{eqn:genPhi}. (b) In practice, we permute the low-resolution Hadamard for better incoherence with the sparsifying wavelet basis. Fast generation of the DSS matrix requires us to impose additional structure on the high-frequency patterns. In particular, each sub-block of the high-frequency pattern is forced to be the same, which enables fast computation via convolutions.}
\label{fig:meas_design}
\end{figure}

We now show the construction of a such a  DSS matrix~$\bPhi$ (see Fig.\ \ref{fig:meas_design}(a)).
A simple way is to start with a $W\times{}W$ Hadamard matrix $\bfH$ and to write the CS matrix as 
\begin{equation}
\bPhi = \bfH \bfD + \bfF,
\label{eqn:genPhi}
\end{equation}
where $\bfD$ is a down-sampling matrix satisfying $\bfD\bfU=\bfI$, and $\bfF \in\reals^{W\times N}$ is an auxiliary matrix that obeys the following constraints: 
\begin{inparaenum}[(i)]
\item The entries of $\bPhi$ are $\pm1$,
\item the matrix~$\bPhi$ has good CS recovery properties (e.g., satisfies the RIP), and
\item $\bfF$ should be chosen such that $\bfF \bfU = \bfZero$.
\end{inparaenum}
Note that an easy way to ensure that $\bPhi$ be $\pm 1$ is to interpret $\bfF$ as sign flips of the Hadamard matrix $\bfH$.
Note that one could chose~$\bfF$ to be an all-zeros matrix; this choice, however, results in a sensing matrix $\bPhi$ having poor CS recovery properties. In particular, such a matrix would inhibit the recovery of high spatial frequencies. 
Choosing random entries in~$\bfF$ such that  $\bfF \bfU = \bfZero$ (i.e., by using random patterns of high spatial frequency) provides excellent performance. 
 
To arrive at an efficient implementation of CS-MUVI, we additionally want to avoid the storage of an entire $W\times{}N$ matrix. To this end, we generate each row $\bff_i \in \reals^N$ of~$\bfF$ as follows:
Associate each row vector $\bff_i$ to an $n \times n$ image of the scene, partition the scene into blocks of size $(n/n_L)\times (n/n_L)$, and associate an $(n/n_L)^2$-dimensional vector $\hat\bff_i$ with each block. We can now use the same vector $\hat\bff_i$ for each block and choose $\hat\bff_i$ such that the full matrix satisfies $\bfF \bfU = \bfZero$. We also permute the columns of the Hadamard matrix $\bfH$ to achieve better incoherence with the sparsifying bases used in \secref{sec:optical} (see Fig.\ \ref{fig:meas_design}(b) for the details). 


\subsection{Preview mode} 
\label{sec:preview}

\begin{figure}[!ttt]
\centering
\includegraphics[width=0.5\columnwidth]{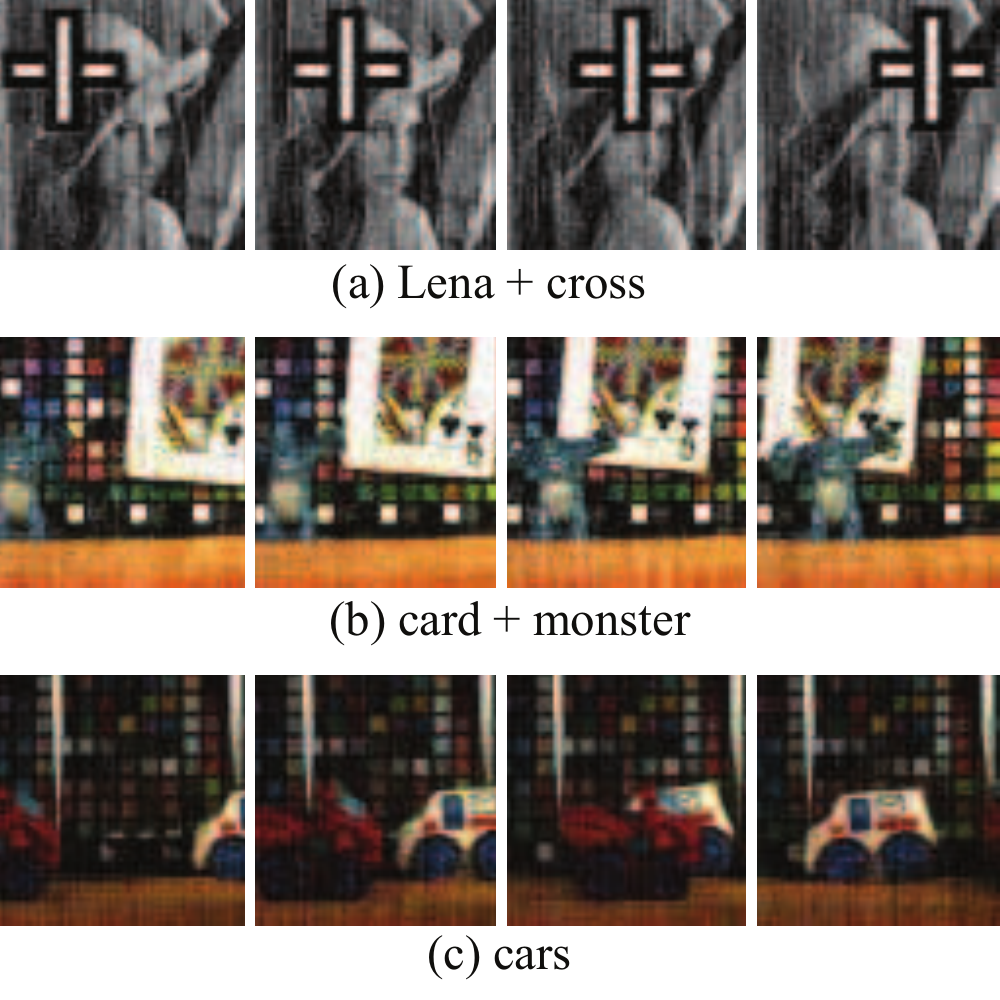}
\caption{\textbf{Preview frames for three different scenes.} All previews consist of $64 \times 64$ pixels. Preview frames are obtained at low computational cost using an inverse Hadamard transform, which opens up a variety of new real-time applications for video CS.}
\label{fig:preview}
\end{figure}

The use of Hadamard matrices for the low-resolution part in the proposed DSS matrices has an additional benefit.
Hadamard matrices have fast inverse transforms, which can significantly speed up the recovery of the low-resolution preview frames.
Such a  ``fast'' DSS matrix has the key capability of generating a high-quality \emph{preview} of the scene (see \figref{fig:preview}) with very low computational complexity; this is beneficial for video CS as it allows one to easily and quickly extract an estimate of the scene motion.
The motion estimate can then be used to recover the video at its full resolution (see  \secref{sec:optical}).
In addition to this, the use of fast DSS matrices can be beneficial in various other ways, including (but not limited to):
%
\paragraph{Digital viewfinder} Conventional SMC architectures do not enable the observation of the scene until CS recovery is performed. Due to the high computational complexity of most existing CS recovery algorithms, there is typically a large latency between the acquisition of a scene and its observation.
Fast DSS matrices offer an \emph{instantaneous} visualization of the scene, i.e., they can provide a real-time digital viewfinder; this capability substantially simplifies the setup of an SMC in practice.

\paragraph{Adaptive sensing} The immediate knowledge of the scene---even at a low resolution---is a key enabler for adaptive sensing strategies. For example, one may seek to extract the changes that occur in a scene from one frame to the next or track the locations of moving objects, while avoiding the typically high latency caused by computationally complex CS recovery algorithms.

\subsection{Selecting $W$}
Crucial to the design of the DSS matrix is the selection of the parameter $W$.
While $W$ is often scene-specific,  a good rule of thumb is as follows: given an $n \times n$ scene, choose $W = n_L^2$ such that the motion of objects is less than $n/n_L$ pixels in the amount of time required to get $W$ measurements. Basically, this would serve to have motion in the preview images restricted to 1 pixel (at the resolution of the preview image).


%


\section{Optical-flow-based video recovery} \label{sec:optical}

We next detail the second part of CS-MUVI, where we obtain the video at a high spatial resolution by estimating and enforcing motion estimates between frames.
%

\subsection{Optical-flow estimation}

Thanks to the preview mode, we can estimate the optical flow between any two (low-resolution) frames $\widehat{\static}_L^i $ and~$\widehat{\static}_L^{j}$.
For CS-MUVI, we compute optical-flow estimates at full spatial resolution between pairs of upsampled preview frames. For the results in the paper, we used ``bicubic'' interpolation to upsample the frames.
This approach turns out to result in more accurate optical-flow estimates compared to an approach that first estimates the optical flow at low resolution followed by upsampling of the optical flow.
Let $\widehat{\static}^i=\bfU \widehat{\static}_L^i$ be the upsampled preview frame. 
The optical flow constraints between two frames, $\widehat{\static}^i$ and $\widehat{\static}^j$,  can be written as
\begin{align*}
\widehat{\static}^i(x, y) = \widehat{\static}^j(x+u_{x, y}, y+v_{x,y}),
\end{align*}
where $\widehat{\static}^i(x, y)$ denotes the pixel $(x,y)$ in the $n\times{}n$ plane of $\widehat{\static}^i$, and  $u_{x,y}$ and $v_{x,y}$ correspond to the translation of the pixel ($x,y$) between frame $i$ and $j$ (see \cite{horn1981determining, liu2009beyond}).

In practice, the estimated optical flow may contain sub-pixel translations, i.e., $u_{x,y}$ and $v_{x,y}$ are not necessarily integer valued. 
If this is the case, then we approximate $\widehat{\static}^j(x+u_{x,y}, y+v_{x,y})$ as a linear combination of its four closest neighboring pixels
\begin{align*}
&\widehat{\static}^j(x+u_{x,y}, y+v_{x,y}) \approx  \sum_{k,\ell\in\{0,1\}} \!\!\! w_{k,\ell}\widehat{\static}^j(\lfloor x+u_{x,y} \rfloor+k, \lfloor y+v_{x,y} \rfloor+\ell),
\end{align*}
where $\lfloor\cdot\rfloor$ denotes rounding towards $-\infty$ and the weights $w_{k,\ell}$ are chosen according to the location within the four neighboring pixels.
In order to obtain robustness against occlusions, we enforce consistency between the forward and backward optical flows; specifically, we discard optical flow constraints at pixels where the sum of the forward and backward flow causes a displacement greater than one pixel.

\subsection{Choosing the recovery frame rate}

Before we detail the individual steps of the CS-MUVI video-recovery procedure, it is important to specify the rate of the frames to be recovered.
When sensing scenes with SMC architectures, there is no obvious notion of frame rate. 
One notion of the frame rate comes from the measurement rate which in the case of the SPC is the operating rate of the DMD.
However, this rate is extremely high and leads to videos whose dimensions are too high to allow feasible computations.
Further, each frame would be associated with a \textit{single} measurement which leads to a severely ill-conditioned inverse problem.
A potential definition comes from the work of Park and Wakin \cite{park2013multiscale} who argue that the frame rate is not necessarily defined by the measurement rate. Specifically, the spatial bandwidth of the video often places an upper-bound on its temporal  bandwidth as well. Intuitively, the idea here is that the larger the pixel size (or smaller the spatial bandwidth), the greater the motion to register a change in the scene.
Hence, given a scene motion in terms of pixels/second, a suitable notion of frame rate is one that ensures sub-pixel motion between consecutive frames.
This notion is more meaningful since it intuitively weaves in the observability of the motion into the definition of the frame-rate.
%
%
Under this definition, we wish to find the largest window size $\Delta W\leq W$ such that there is virtually no motion at full  resolution ($n \times n$). 
%
In practice, an estimate of $\Delta W$ can be obtained by analyzing the preview frames.
Hence, given a total number of $T$ compressive measurements,  we ultimately recover $F = T/\Delta W$ full-resolution  frames.
Note that a smaller value of $\Delta W$ would decrease the amount of motion associated with each recovered frame;  this would, however, increase the computational complexity (and memory requirements) substantially as the number of full-resolution frames to be recovered increases. 
Finally, the choice of $\Delta W$ is inherently scene-specific; scenes with fast moving highly textured objects require a smaller $\Delta W$ as compared to those with slow moving smooth objects.
The choice of $\Delta W$ could potentially be made time-varying as well and derived from the preview; this showcases the versatility of having the preview and is an important avenue for future research.
%
\subsection{Recovery of full-resolution frames}

We are now ready to detail the final stage of CS-MUVI.
Assume that $\Delta W$ is chosen such that there is little to no motion associated with each preview frame.
Next, associate a preview frame with a high-resolution frame ${\bfx}_k$, $k\in\{1,\ldots,T\}$ by grouping $W=N_L$ compressive measurements in the immediate vicinity of the frame (since $\Delta W \le W$).
Then, compute the optical-flow between successive (up-scaled) preview frames.

We can now recover the  high-resolution video frames as follows.
We enforce sparse spatio-temporal gradients using the 3D total variation (TV) norm.
We furthermore consider the following two constraints: 
\begin{inparaenum}[(i)]
\item Consistency with the acquired CS measurements, i.e, $\bfy_t = \inner{\phi_t}{\bfx_{I(t)}}$, where $I(t)$ maps the sample index~$t$ to the associated frame index~$k$, and
\item estimated optical-flow constraints between consecutive frames.
\end{inparaenum}
Together, we arrive at the following convex optimization problem:
\begin{align*}
\text{(TV)} \, \left\{\begin{array}{ll}
\!\text{minimize} & \text{TV}_\text{3D}(\bfx) \\[0.2cm]
\!\text{subject to} & \!\!\!  \left\| \inner{\phi_t}{\bfx_{I(t)}} - y_t  \right \|_2 \le \epsilon_1,  \\
& \!\!\! \left\|  \bfx_i(x, y) \!-\! \bfx_j(x+u_x, y+ v_y) \right\|_2 \le \epsilon_2, 
\end{array}\right.
\end{align*}

which can be solved using standard convex-optimization techniques.
The specific technique that we employed was by variable splitting and using ALM/ADMM.

The parameters $\epsilon_1$ and $\epsilon_2$ are indicative of the measurement noise levels and the inaccuracies in the brightness constancy, respectively.
$\epsilon_1$ captures all sources of measurement noise including photon, dark, and read noise.
Photon noise is signal dependent. However, in an SPC, each measurement is the sum of a random selection of half the micromirrors on the DMD. For most natural scenes, we can expect the measurements to be tightly clustered --- to be more specific, around one-half of the total light-level of the scene. Hence, the photon noise will have nearly the same variance across the measurements.  
Hence, for the SPC, all sources of measurement noise can be clubbed into one parameter $\epsilon_1$ which   is set via a calibration process.
Setting $\epsilon_2$ is based on the thresholds used in detecting violation of brightness constancy when estimating brightness constancy. 
For the results in this paper, $\epsilon_2$ is set to $0.02 \times \sqrt{P}$, where $P$ is the total number of pixel pairs for which we enforce brightness constancy.
%

%


\section{Evaluation and Comparisons} \label{sec:experiments}

In this section, we validate the performance and capabilities of the CS-MUVI framework using simulations. Results on real data obtained from our SPC lab prototype are presented in \secref{sec:real}.
All simulation results were generated from high-speed videos  having a spatial resolution of $n\times n =256 \times 256$ pixels. The preview videos have a spatial resolution of  $64 \times 64$ pixels with (i.e., $W=4096$). 
We assume an SPC architecture as described in \cite{duarte2008single} with parameters chosen to mimic operation of our lab setup.
Noise was added to the compressive measurements using an i.i.d.\ Gaussian noise model such that the resulting SNR was 60\,dB. 
Optical-flow estimates were extracted using the method described in \cite{liu2009beyond}.
The computation time of CS-MUVI is dominated by both optical flow estimation and solving $(TV)$. Typical runtimes for the entire algorithm are  2--3 hours on an off-the-shelf quad-core CPU for a video of resolution $256\times256$ pixels with $256$ frames. However, computation of the low-resolution preview can be done almost instantaneously. 

\paragraph{Video sequences from a high-speed camera}

\begin{figure*}[ttt]
\centering
\includegraphics[width=\textwidth]{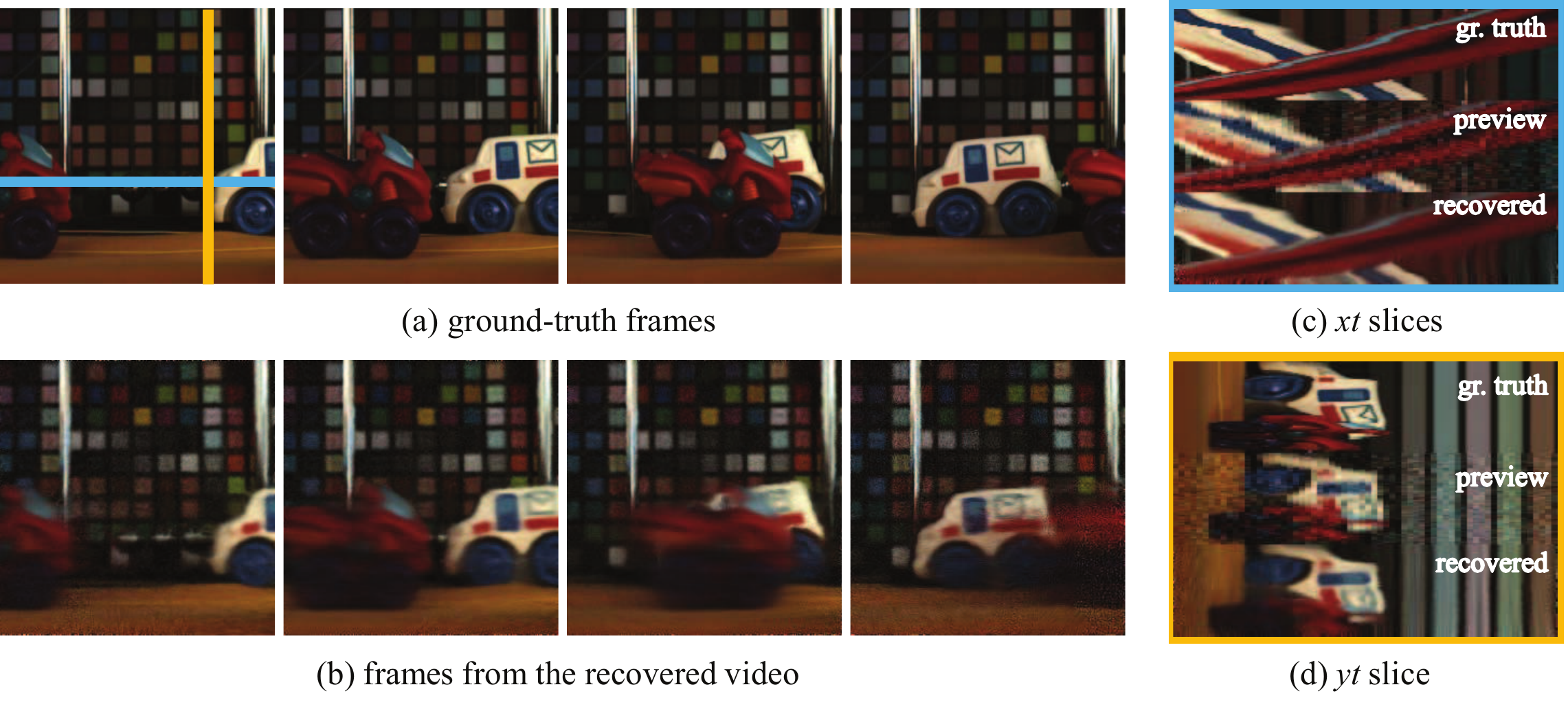}
\caption{\textbf{Recovery on high-speed videos.} CS-MUVI recovery results of a video obtained from a high-speed camera operating at $250$ fps. Shown are frames of (a) the ground truth and~(b) the recovered video (PSNR = $25.0$\,dB). The $xt$ and $yt$ slices shown in (c) and (d) correspond to the color-coded lines of the first frame in~(a). Preview frames for this video are shown in Fig.\ \ref{fig:preview}. (The $xt$ and $yt$ slices are rotated clockwise by 90 degrees.)}
\label{fig:carcar}
\end{figure*}

\begin{figure*}[ttt]
\centering
\includegraphics[width=\textwidth]{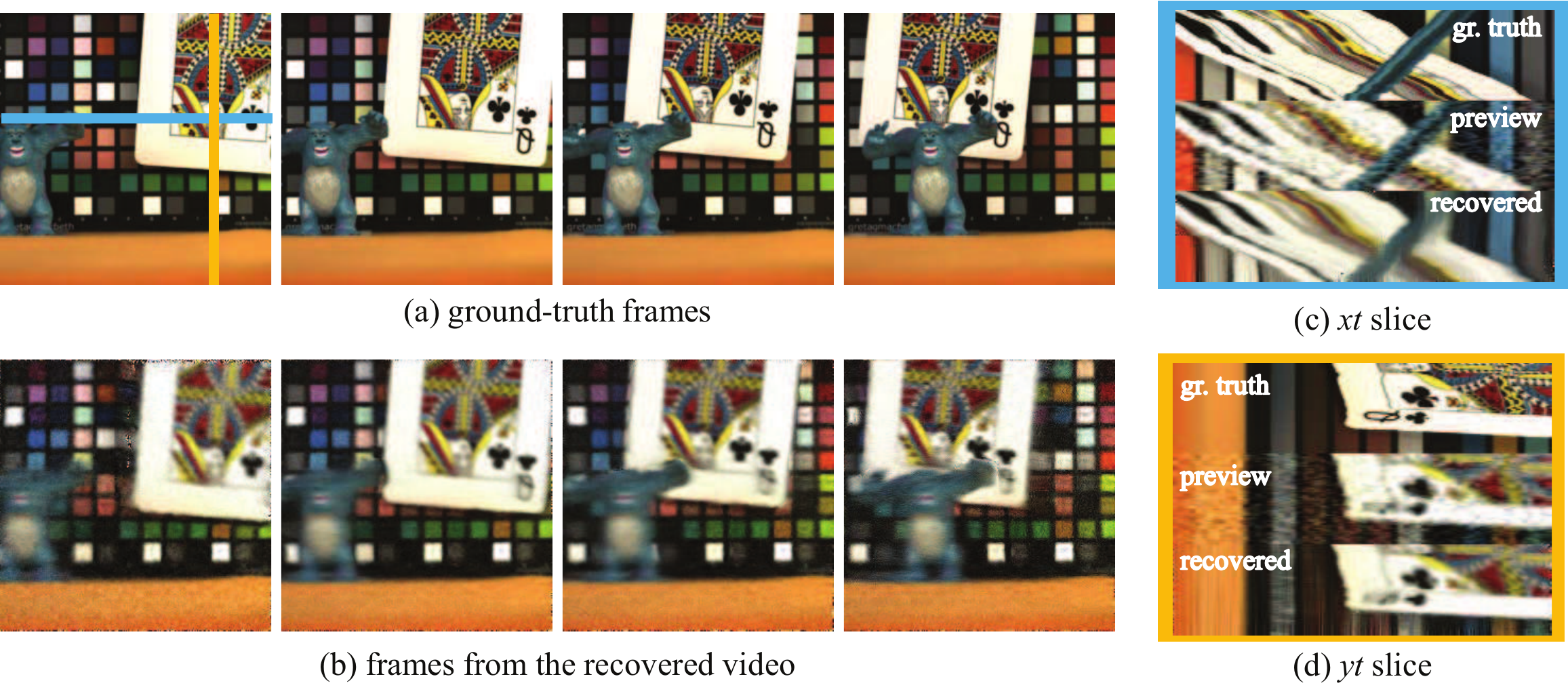}
\caption{\textbf{Recovery on high-speed videos.} CS-MUVI recovery results of a video obtained from a high-speed camera. Shown are frames of (a) the ground truth and~(b) the recovered video (PSNR = $20.4$\,dB). The $xt$ and $yt$ slices shown in (c) and (d) correspond to the color-coded lines of the first frame in~(a). Preview frames for this video are shown in Fig.\ \ref{fig:preview}. (The $xt$ and $yt$ slices are rotated clockwise by 90 degrees.)}
\label{fig:cardmonster}
\end{figure*}

The results shown in Figs.~\ref{fig:carcar} and \ref{fig:cardmonster} correspond to scenes acquired by a high-speed (HS) video camera operating at 250 frames per second.
Both videos show complex (and fast) movement of large objects as well as severe occlusions.
For both sequences, we emulate an SPC operating at $8192$  compressive measurements per second.
For each video, we used $2048$ frames of the HS camera to obtain a total of $T=32 \times 2048$ compressive 
measurements. The final recovered video sequences consist of $F=61$ frames $(\Delta W = 1024)$.
Both recovered videos demonstrate the effectiveness of CS-MUVI.


\begin{figure*}[ttt]
\centering
\includegraphics[width=\textwidth]{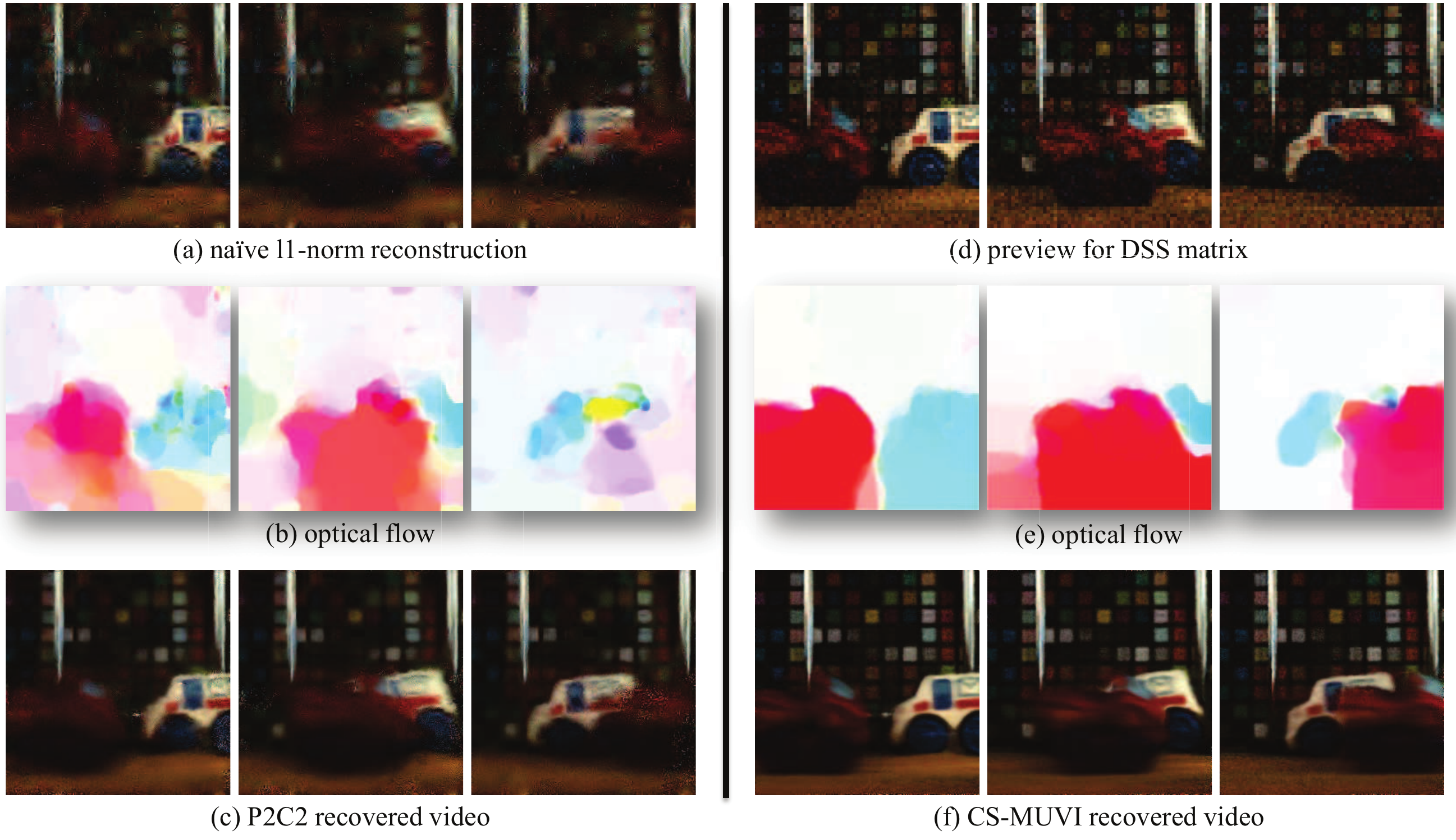}
\caption{\textbf{Comparisons to the two-step strategy used in the P2C2 camera \cite{reddy2011p2c2}.} Shown are frames of (a) reconstruction obtained by minimizing the $\ell_1$-norm of wavelet coefficients, (b) the resulting optical-flow estimates, and (c) the P2C2 recovered video. The frames in (d) correspond to preview frames when using DSS matrices, (e) are the optical-flow estimates, and (f) is the scene recovered by CS-MUVI.}
\label{fig:comparisonwithp2c2}
\end{figure*}

\paragraph{Comparison with the P2C2 algorithm}
In the P2C2 camera \cite{reddy2011p2c2}, a two-step recovery algorithm --- similar to CS-MUVI --- is presented. This algorithm is near-identical to CS-MUVI except that the measurement model does not use DSS measurement matrices; hence, an initial recovery using wavelet sparse models is  used to obtain an initial estimate that plays the role of the preview frames.
Figure~\ref{fig:comparisonwithp2c2} presents the results of both CS-MUVI and the recovery algorithm for the P2C2 camera \cite{reddy2011p2c2}, with the same number of measurements/compression level.
It should be noted that the P2C2 camera algorithm was developed for temporal multiplexing cameras and  \emph{not} for SMC architectures. 
Nevertheless, we observe from Figs.~\ref{fig:comparisonwithp2c2} (a) and~(d) that na\"ive $\ell_1$-norm recovery delivers significantly worse initial estimates than the preview mode of CS-MUVI. The advantage of CS-MUVI for SMC architectures is also visible in the corresponding optical-flow estimates (see Figs.~\ref{fig:comparisonwithp2c2} (b) and (e)). The P2C2 recovery algorithm has substantial artifacts, whereas the result of CS-MUVI is visually pleasing. 
In all, this demonstrates the importance of the DSS  matrix and the ability to robustly obtain a preview of the video.

\begin{figure*}[ttt]
\centering
\includegraphics[width=\textwidth]{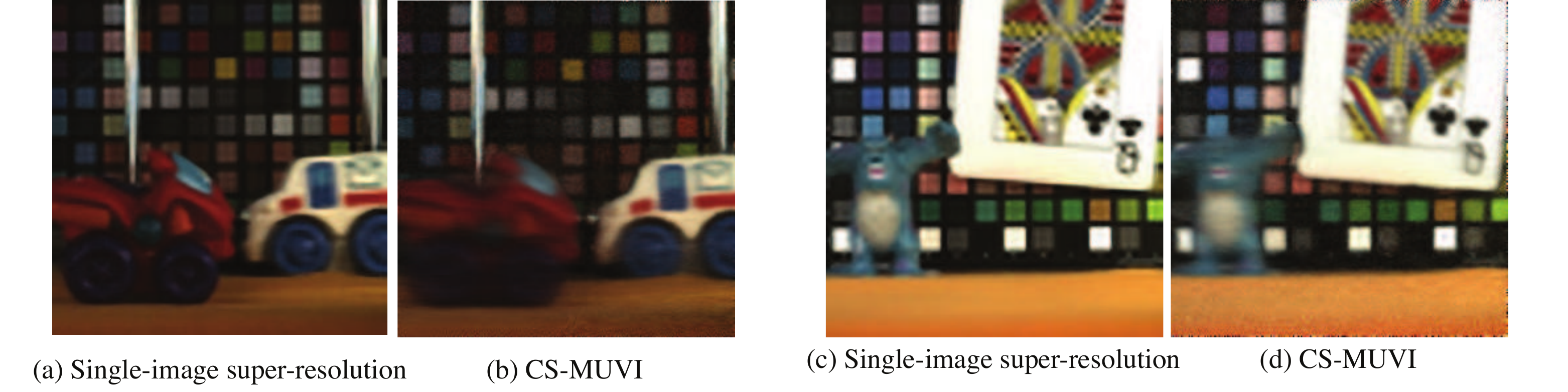}
\caption{\textbf{Comparisons to the single-image super-resolution algorithm of \cite{yang2012coupled}.} Shown are results  on two high-speed videos. (a,c) We use a low-resolution Hadamard matrix to sense a low-resolution image with $64\times 64$ pixels and subsequently, super-resolve them $4\times$. (b, d)We use DSS matrices instead of low-resolution Hadamard to obtain the CS-MUVI results. Both algorithms have the same measurement rate. We observe that performance of CS-MUVI is similar to that of the super-resolution algorithm.}
\label{fig:sr}
\end{figure*}

\begin{figure*}[ttt]
\centering
\includegraphics[width=\textwidth]{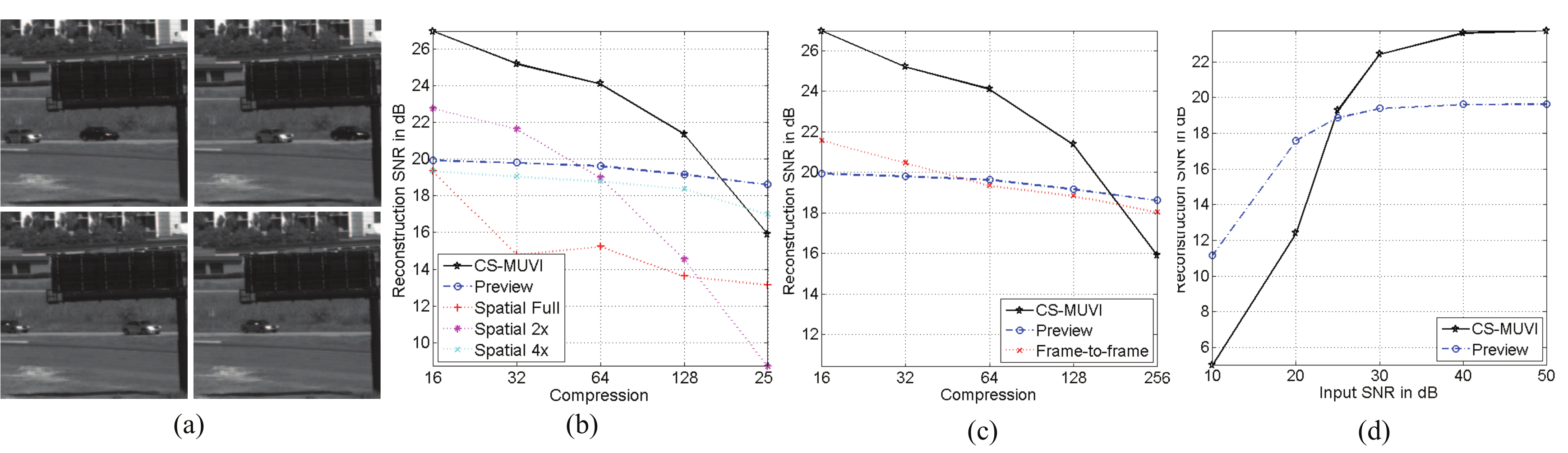}
\caption{\textbf{Quantitative performance.} (a) Four frames from a high-speed video. (b) Performance of CS-MUVI for different compression ratios compared against ``Nyquist'' cameras that trade-off spatial and temporal resolution to achieve the desired compression.
(c) Performance of CS-MUVI compared against video recovered using frame-to-frame sparse wavelet prior. For the sparse wavelet prior, for each compression ratio, the window of measurements associated with each recovered frame was varied and the best performing result is shown.
(d) Performance of CS-MUVI for varying levels of AWGN. For high noise levels (low input SNR), the low quality preview leads to poor optical flow estimates which causes a severe degradation in performance.}
\label{fig:performance}
\end{figure*}

\paragraph{Comparisons against single-image super-resolution}
There has been remarkable progress in single image super-resolution (SR). 
Figure \ref{fig:sr} compares CS-MUVI to a sparse dictionary-based super-resolution algorithm \cite{yang2012coupled}.
From our observations, the results produced by the super-resolution  are comparable to CS-MUVI when the upsampling is about $4 \times$. 
However, in spite of this, the best known results in SR seldom produce meaningful results beyond $4\times$ super-resolution. 
Our proposed technique is in many ways similar to SR except that we obtain multiple coded measurements of the scene and this allows us to obtain higher super-resolution factors at potential loss in temporal resolution.

\paragraph{Performance analysis}
Finally, we look at quantitative evaluation of CS-MUVI for varying compression ratios and input measurement noise level. Our metric for performance is reconstruction SNR in dB defined as follows:
\[ \textit{RSNR} = -20 \log_{10}\left( \frac{\| \bfx - \widehat{\bfx} \|_2}{\| \bfx \|_2}\right), \]
where $\bfx$ and $\widehat{\bfx}$ are the ground truth and estimated video, respectively.
The test-data for this is a 250 fps video of vehicles on a highway. A few frames from this video are shown in \figref{fig:performance}(a).
We establish a baseline for these results using two different algorithms.
First, we consider ``Nyquist  cameras'' that blindly tradeoff spatial and temporal resolution to achieve the desired compression. For example, at a compression factor of $16\times$, a Nyquist camera could deliver full-resolution at $1/16$-th the temporal resolution or deliver $1/2$-th the spatial resolution at $1/8$-th the temporal resolution, and so on. This spatio-temporal trade off is feasible in most traditional imagers by binning pixels at readout.
Second, we consider videos recovered using na\"ive frame-to-frame wavelet priors. For such reconstructions, we optimized over different window lengths of measurements associated with each recovered frame and chose the setting that provided the best results.
Figure \ref{fig:performance}(b,c) show reconstruction SNR for CS-MUVI and the two baseline algorithms for varying levels of compression.
At high compression ratios,  the performance of CS-MUVI suffers from poor optical-flow estimates.
Finally, in \figref{fig:performance}(d), we present performance for varying level of measurement or input noise. Again, as before, for high noise levels, optical flow estimates suffer leading to poorer reconstructions.
In all, CS-MUVI delivers high quality reconstructions for a wide range of compression and noise levels.

\begin{figure}[!hhh]
\center
\includegraphics[width=\textwidth]{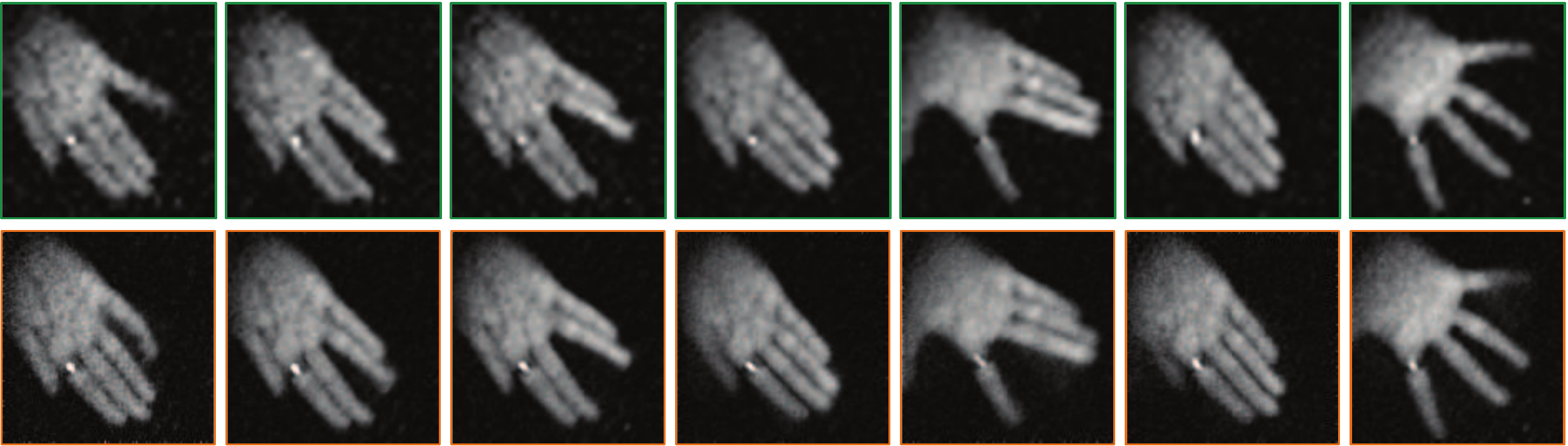} \\
(a) Hand: Simple motion \\
\includegraphics[width=\textwidth]{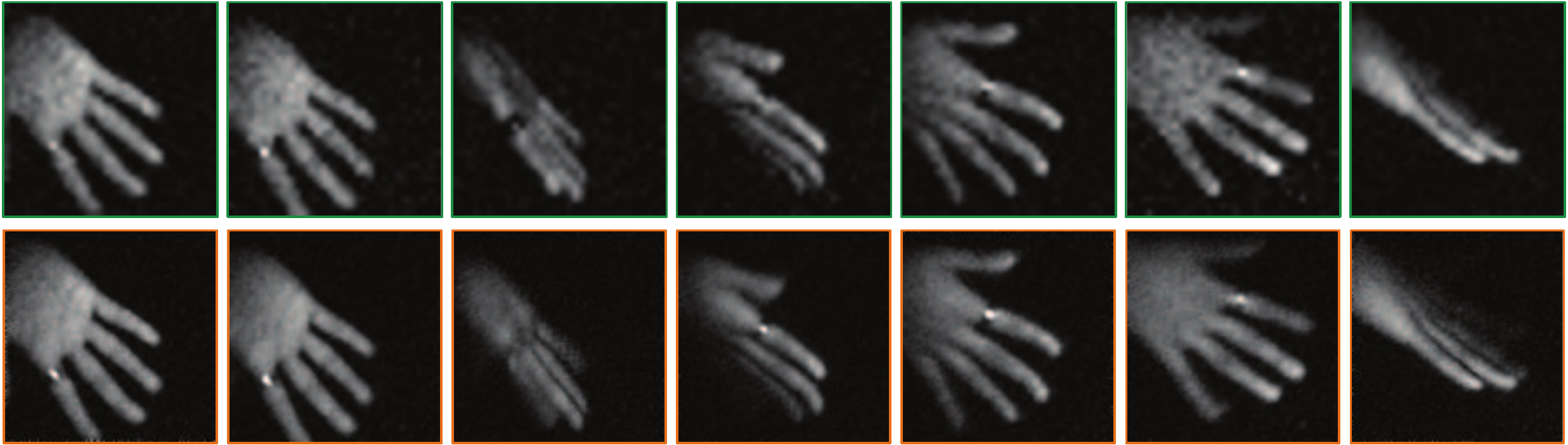}\\
(b) Hand: Complex motion \\
\includegraphics[width=\textwidth]{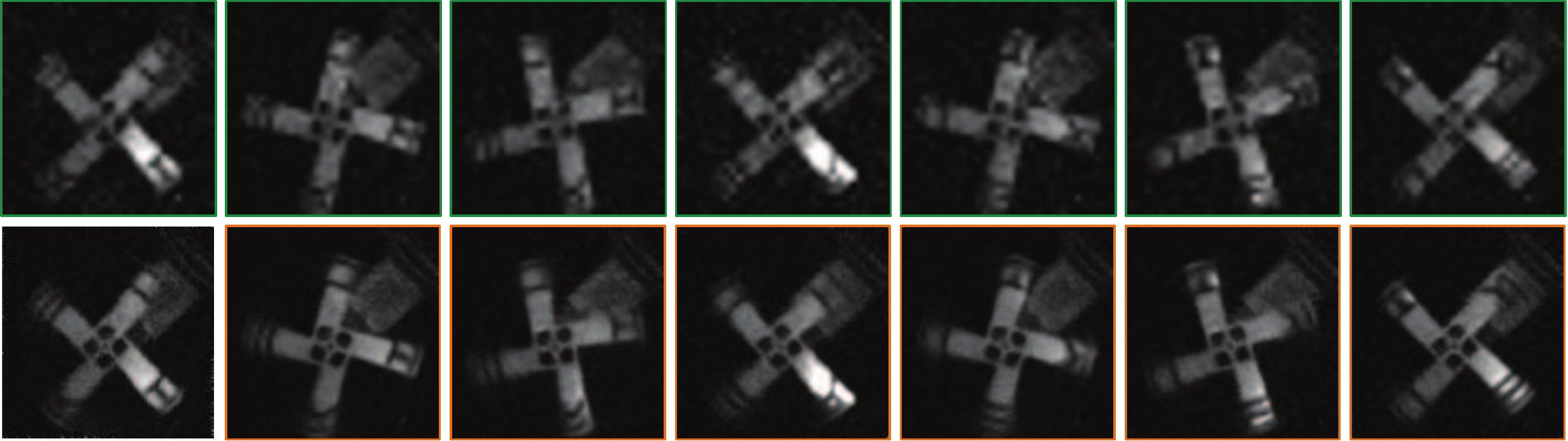}\\
(c) Windmill \\
\includegraphics[width=\textwidth]{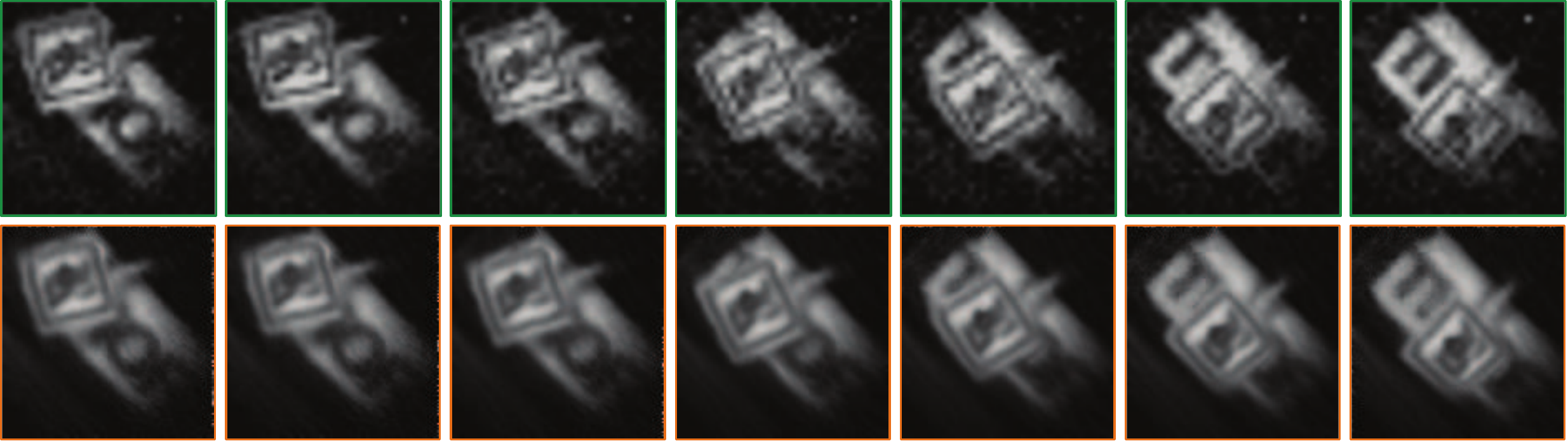}\\
(d) Pendulum
\caption{\textbf{Reconstructions from SPC hardware.} Shown above (a-d) are four different scenes with different kinds of motion. For each scene, the top row (marked in green) shows frames from the preview, and the bottom row (red) shows the corresponding frames from the final recovered video.}
\label{fig:mosaic}
\end{figure}

\section{Hardware implementation} \label{sec:real}

We now present video recovery results on real data from our SPC lab prototype.

\paragraph{Hardware prototype} 
The SPC setup we used to image real scenes is comprised of a DMD operating at 10,000 mirror-flips per second.
The real measured data was acquired using a SWIR photodetector for the scenes involving the pendulum and a  visible photodetector for the rest (the hand and windmill scene).
While the DMD we used is capable of imaging the scene at a XGA resolution (i.e., 1024$\times$768 pixels),
we operate it at a lower spatial resolution mainly, for two reasons.
First, recall that the measurement bandwidth of an SPC is determined by the speed of operation of the DMD. In our case, this  was 10,000 measurements per second. Even if we were to obtain a compression of $50\times$, then our device would be similar to a conventional sampler whose measurement bandwidth is $5\times10^5$ measurements/sec which would result in a video of approximately $128\times 128$ pixels at 30 frames/sec. Hence, we operate it at a spatial resolution of $128 \times 128$ pixels by grouping pixels together on the DMD  as one $6 \times 6$  super-pixel. 
Second,  the patterns displayed on the DMD were required  to be preloading onto the memory board attached to DMD via a USB port. With limited memory, typically 96 GB, any
reasonable temporal resolution with XGA resolution would be infeasible on our current SPC prototype.
We emphasize that both of these are limitations due to the used prototype and not of the underlying algorithms. Recent, commercial DMDs can operate at least $1$-to-$2$ orders of magnitude faster \cite{narasimhan2008temporal} and the increase in measurement bandwidth would enable sensing at higher spatial and temporal resolutions.

\paragraph{Gallery of real data results}
Figure \ref{fig:mosaic} shows a few example reconstructions from our SPC lab setup.
Each video is approximately $1.6$ seconds long and correspond to $M=16384$ measurements from the SPC.
With $D = 4$, all previews (the top row in each sub-image in \ref{fig:mosaic}) were each of size  $32\times32$ pixels. Videos were recovered with $F = 125$ frames.
The supplemental material has videos for each of the results.

%
%

\begin{figure}[!ttt]
\center
\includegraphics[width=\textwidth]{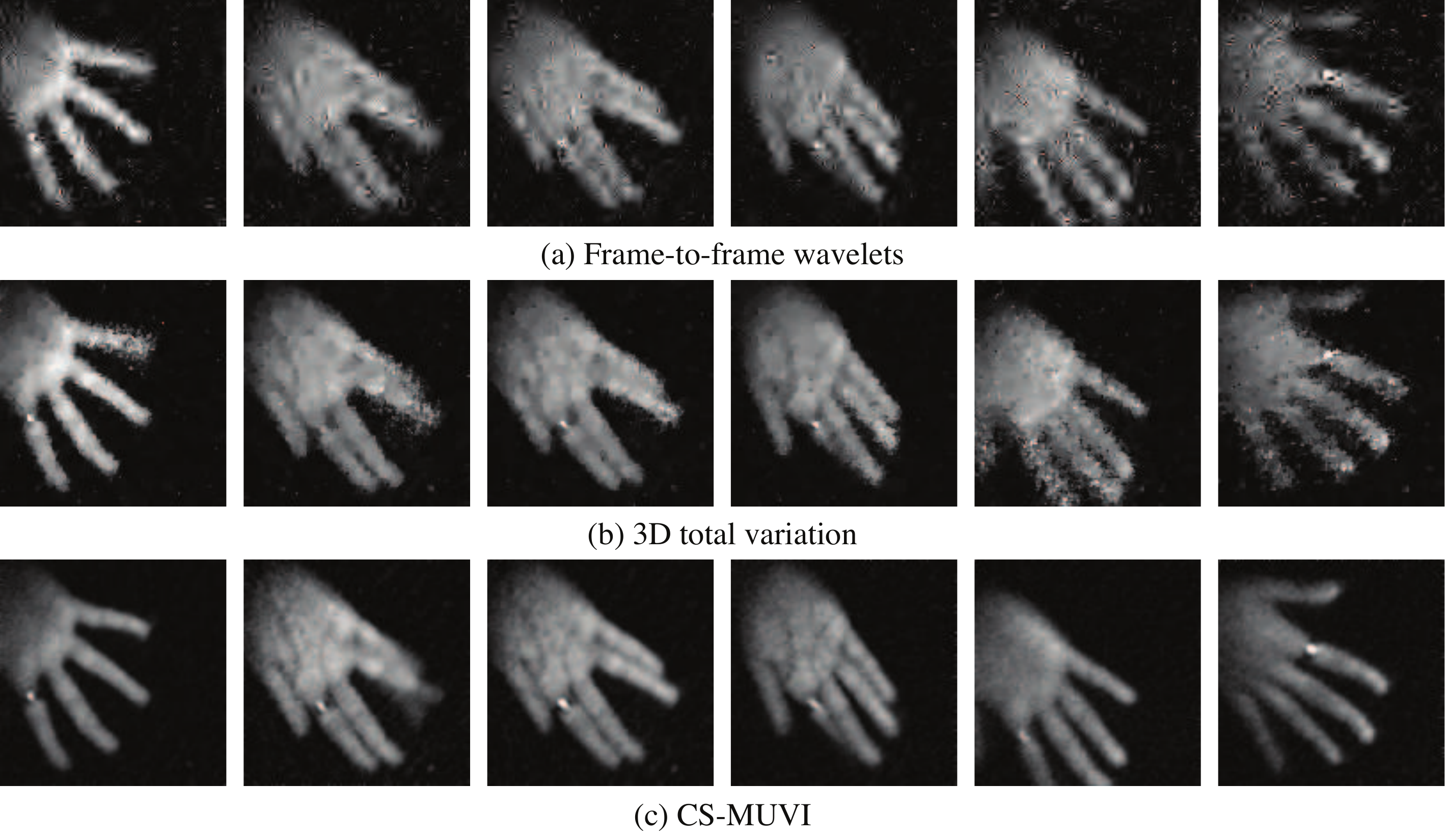}
\caption{\textbf{Performance comparison of different signal models.} We look at performance of various signal models for a dynamic, fast moving target. Shown are select frames where the speed of the target was high. As before, CS-MUVI handles fast moving targets gracefully without any of the artifacts present in competing signal models. Refer to the supplemental material for a complete video.}
\label{fig:priors}
\end{figure}

\begin{figure}[!ttt]
\center
\includegraphics[width=\textwidth]{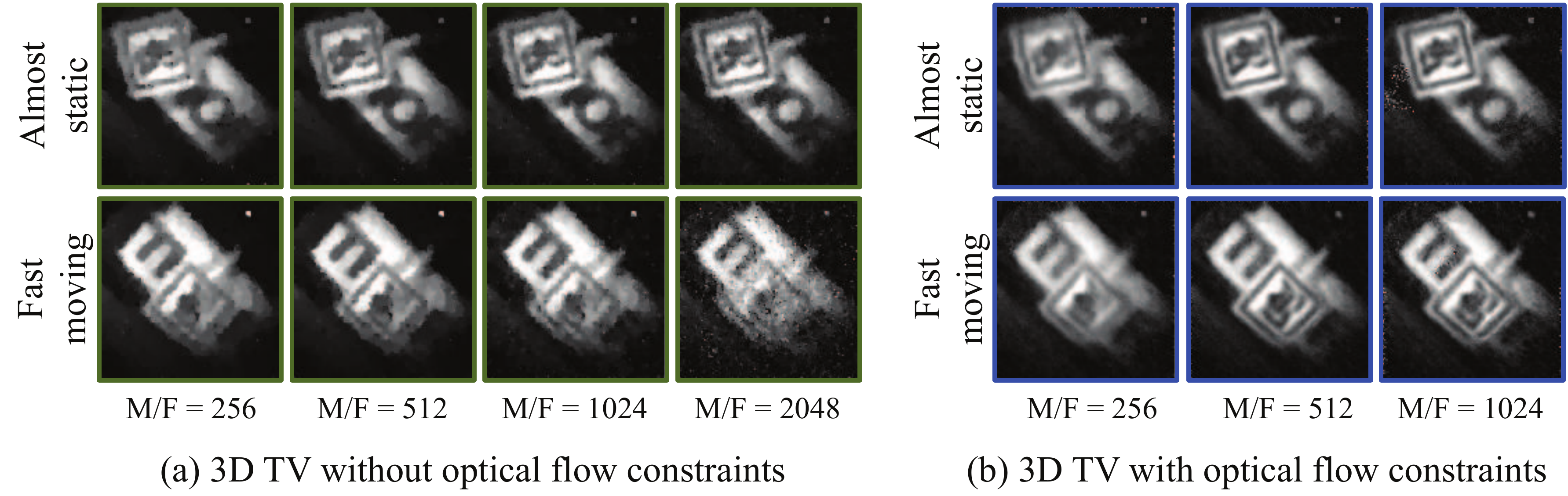}
\caption{\textbf{Comparison of recovered videos with and without optical-flow constraints.} 
Data was collected with a SPC operating at 10,000\,Hz with a SWIR photodetector.
A total of $M=16,384$ compressive measurements were obtained at a DMD resolution of $128 \times 128$.
In each case, we show multiple reconstructions with different number of compressive measurements associated with each frame. That is, in each instance, the number of recovered frame $F$ is chosen to satisfy the target $M/F$ value.
(a) Reconstructions without optical flow constraints. The top row shows the pendulum at one end of its swing where it is nearly stationary. The bottom row shows the pendulum when it is moving the fastest. As expected, increasing the number of measurements per frame, $M/F$, increases the motion blur significantly.
(b) In contrast, use of optical flow preserves the quality of results. The visual quality peaks at $M/F = 512$ (see supplemental videos).}
\label{fig:With_Without_OF}
\end{figure}

\paragraph{Role of different signal priors}
Figures \ref{fig:signalmodels}, \ref{fig:priors}, and \ref{fig:With_Without_OF} show the performance of three different signal priors on the same set of measurements.
In \figref{fig:signalmodels}, we compare wavelet sparsity of the individual frames, 3D total variation, and CS-MUVI, which uses optical flow constraints in addition to the 3D total variation model. CS-MUVI delivers superior performance in recovery of the spatial statistics (the textures on the individual frames) as well as temporal statistics (the textures on temporal slices).
In \figref{fig:priors}, we look at specific frames across a wide gamut of reconstructions where the target motion is very high. Again, we observe that reconstructions from CS-MUVI is not just free from artifacts, it also resolves spatial features better (ring on the hand, palm lines, etc.).
Finally, for completeness, in \figref{fig:With_Without_OF}, we vary the number of measurements associated with each frame for both 3D total variation and CS-MUVI.
Predictably, while the performance of 3D total variation is poor for fast moving objects, CS-MUVI delivers high-quality reconstructions across a wide range of target motion.

\begin{figure}[!ttt]
\center
\includegraphics[width=\textwidth]{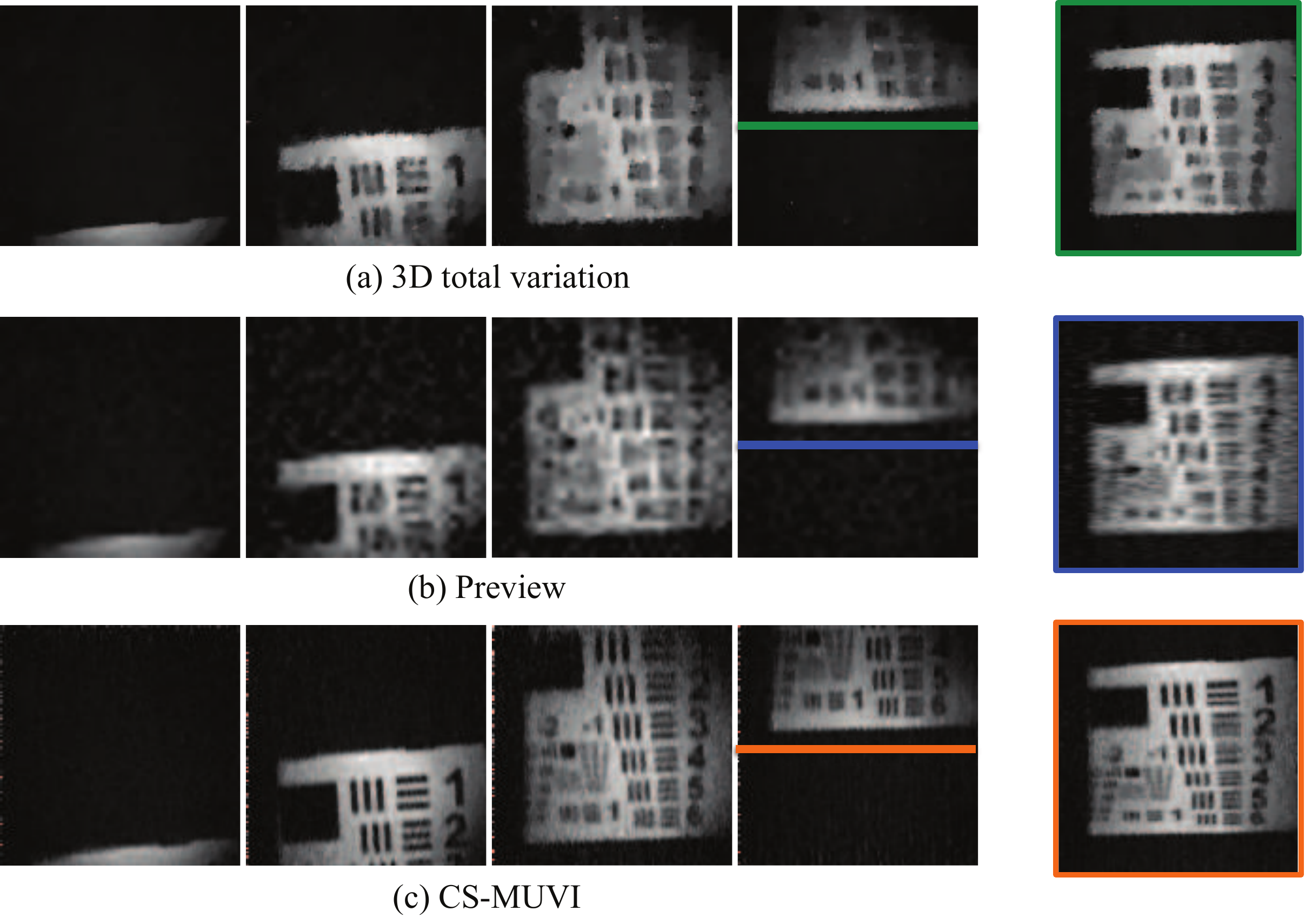}
\caption{\textbf{Resolution chart.} Reconstruction results on a translating resolution chart at a compression ratio of $32\times$.
In each row, we show frames from the recovered video as well as its xt slice in the color coded box in the last column.
}
\label{fig:reschart}
\end{figure}

\begin{figure}[!ttt]
\center
\includegraphics[width=\textwidth]{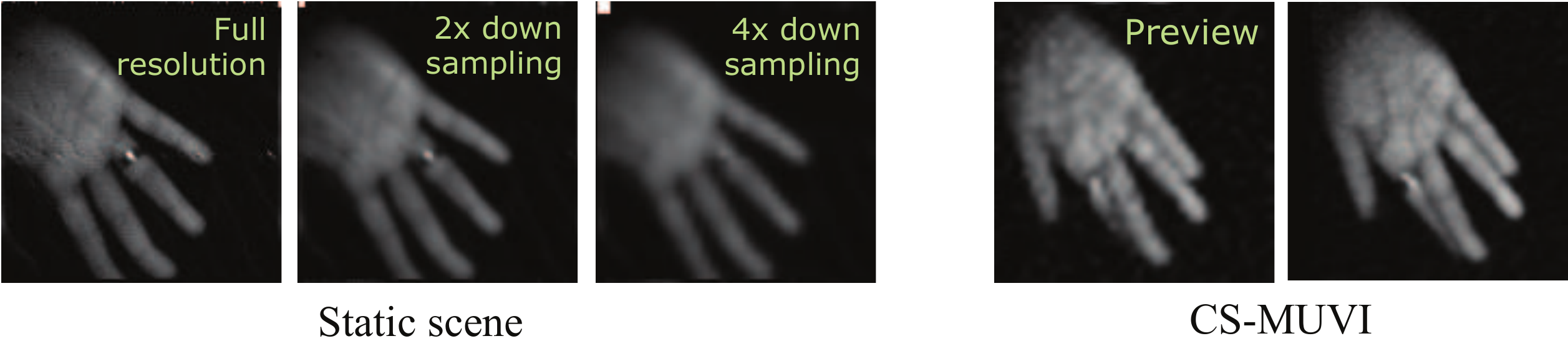}
\caption{\textbf{Achieved resolution.} We compare the achieved spatial resolution of the recovered video for a static target. For visual comparison, we artificially downsample the static image. It is clear that CS-MUVI recovers spatial resolution higher than a 2$\times$ downsampling but slightly worse than the full resolution  image.}
\label{fig:rescmp}
\end{figure}

\paragraph{Achieved spatial resolution.}
In \figref{fig:reschart} and \figref{fig:rescmp} , 
Note that a SMC seeks to super-resolve a low resolution sensor using optical coding and spatial light modulators. Hence, it is of utmost importance to verify if the device actually delivers on the promised improvement in spatial resolution. 

In \figref{fig:reschart}, we present reconstruction results on a  resolution chart.
The resolution chart was translated so as to enter and exit the field-of-view of the SPC within 8 seconds providing a total of $86000$ measurements. A video with $159$ frames was recovered from these measurements for an overall compression ratio of $32\times$. 
\figref{fig:reschart} indicates that the CS-MUVI recovers spatial detail to a per-pixel precision validating the claims of achieved compression. 
For this result, we regularized the optical flow to be translational. Specifically, after estimating the flow between the preview frames, we used the median of the flow-vectors as a global translational flow.

In \figref{fig:rescmp}, we characterize the spatial resolution achieved by CS-MUVI by comparing it to the image of a static scene obtained using pure Hadamard multiplexing.
As expected, we observe that the preview image is  the same resolution as the static image downsampled $4\times$. 
Frames recovered from CS-MUVI exhibit sharper texture than a $2\times$ downsampling of the static frame, but slightly worse than the full-resolution static image. Note that this scene contained complex non-rigid and fast motion.

\begin{figure}[!ttt]
\center
\includegraphics[width=\textwidth]{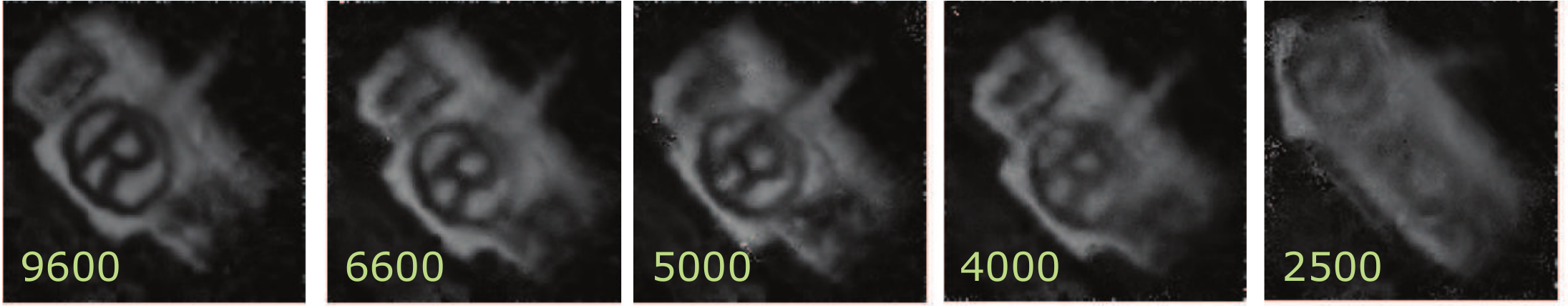}
\caption{\textbf{Performance for varying speed.} We slowed down the operating speed of the SPC to indirectly increase object speed. The operating speed of the SPC is overlaid on top of the recovered video. 
Shown is a single frame from each recovered video; the instant corresponding to the pendulum swinging at maximum speed.}
\label{fig:speedo}
\end{figure}

\begin{figure}[!ttt]
\center
\includegraphics[width=\textwidth]{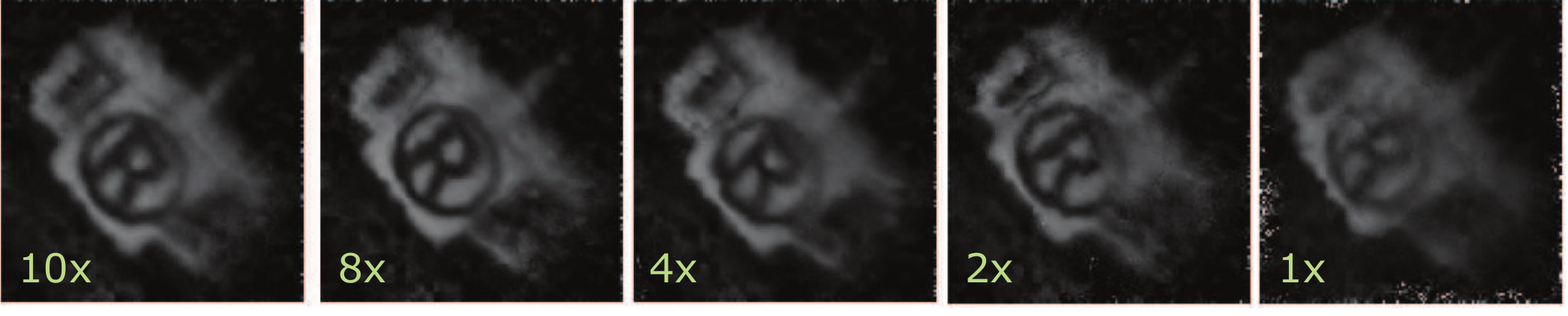}
\caption{\textbf{Performance  for varying scene illumination levels.} We controlled the total light level in the scene by controlling the light throughput of the illumination sources. Shown above are results at different scene light levels---each case calibrated by the multiple of the minimum light level. In each case, we show one frame of the recovered video; the instant corresponding to the pendulum swinging at maximum speed. The performance degradation of the algorithm is graceful with only little artifacts.}
\label{fig:illum}
\end{figure}

\begin{figure}[!ttt]
\center
\includegraphics[width=\textwidth]{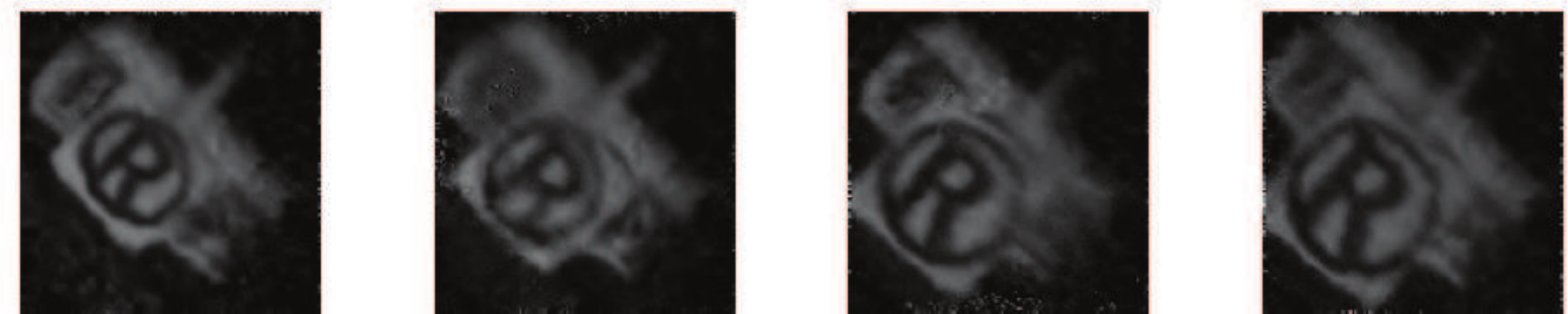}
\caption{\textbf{Performance for varying size of dynamic object.} For a wide range of object size, ranging from a quarter to half of the entire field-of-view of the camera, we obtain stable reconstructions.}
\label{fig:size}
\end{figure}

\paragraph{Variations in speed, illumination, and size}
Finally, we look at performance on real data for varying levels of scene illumination, object speed and size.
For  illumination (\figref{fig:illum}), we use the SPC measurement level as a guide to the amount of scene illumination.
For object speed (\figref{fig:speedo}), we instead slow down the DMD since it indirectly provides finer control on the apparent speed of the object.
For size (\figref{fig:size}), we vary the size of the moving target.
In all cases, we show the recovered frame corresponding to the object moving at the fastest speed.
The performance of CS-MUVI degrades gracefully across all variations. The interested reader is referred to supplemental material for videos of these results.

\section{Discussion} \label{sec:discuss}

\sloppy
\paragraph{Summary} 
The promise of an SMC is to deliver high spatial resolution images and videos from a low-resolution sensor.
The most extreme form of such SMCs is the SPC which poses a single photodetector or a sensor with no resolution by itself.
In this paper, we demonstrate---for the very first time on real data---successful video recovery at $128\times$ super-resolution for fast-moving scenes.
This result has important implications for regimes where high-resolution sensors are prohibitively expensive.
A example of this is imaging in SWIR; to this end, we show results using a SPC with a photodetector tuned to this spectral band.

At the heart of our proposed framework  is the design of a novel class of sensing matrices and an optical-flow based video reconstruction algorithm.
In particular, we have proposed dual-scale sensing (DSS) matrices that 
\begin{inparaenum}[(i)]
\item exhibit no noise enhancement when performing least-squares estimation at low spatial resolution and 
\item preserve information about high spatial frequencies.
\end{inparaenum}
We have developed a DSS matrix having a fast transform, which enables us to compute instantaneous {\em preview} images of the scene at low cost.
The preview computation supports a large number of novel applications for SMC-based devices, such as providing a digital viewfinder, enabling human-camera interaction, or triggering adaptive sensing strategies.


\paragraph{Limitations}
Since CS-MUVI relies on optical-flow estimates obtained from low-resolution  images, it can fail to recover  small objects with rapid motion.
More specifically, moving objects that are of sub-pixel size in the preview mode are lost. 
Figure \ref{fig:carcar} shows an example of this limitation: The cars are moved using fine strings, which are visible in \figref{fig:carcar}(a)  but not in \figref{fig:carcar}(b).
Increasing the spatial resolution of the preview images eliminates this problem at the cost of more motion blur. To avoid these limitations altogether, one must increase the sampling rate of the SMC.
In addition, reducing the complexity of solving (PV) is of paramount importance for practical implementations of CS-MUVI.

\paragraph{Faster implementations}
Current implementation of CS-MUVI take in the order of hours for high-resolution videos with a large number of frames.
This large run-time can be attributed to the DSS matrix lacking a fast transform as well as the inherent complexity associated with high-resolution signals.
Faster implementations of the recovery algorithm is an interesting research directions.

\paragraph{Multi-scale preview}
A drawback of our approach is the need to specify the resolution at which preview frames are
recovered; this requires prior knowledge of object speed.
An important direction for future work is to relax this requirement via the construction of
multi-scale sensing matrices that go beyond the DSS matrices proposed here.
The recently proposed sum-to-one (short STOne) transform \cite{goldstein2013stone} provides such a multi-scale sensing matrix. 
Specifically, the STOne transform is a carefully designed Hadamard transform that remains a Hadamard transform of a lower-resolution when downsampled.
Using the STOne transform in place of the DSS matrix could potentially provide previews of various spatial resolutions.

\paragraph{Multi-frame optical flow}
The majority of the artifacts in the reconstructions stem from inaccurate optical-flow estimates---a result of residual noise in the preview images.
It is worth noting, however, that we are using an off-the-shelf optical-flow estimation algorithm; such an approach ignores the continuity of motion across \emph{multiple} frames. We envision significant performance improvements if we use multi-frame optical-flow estimation \cite{rubinstein2012towards}.
Such an approach could potentially alleviate some of the challenges faced in pairwise optical flow including the inability to recover precise flow estimates for both slow-moving and fast-moving targets.
%
%

\paragraph{Towards high-resolution imagers}
The spatial resolution of an SMC is limited by the resolution of the spatial light modulator. Commercially available DMDs, LCDs and LCoSs have a spatial resolution of $1$--$2$ megapixels. 
An important direction for future research is the design of imaging architectures, signal models and recovery algorithms to obtain videos at this spatial resolution (and say, 30 fps temporal resolution). The key stumbling block for an SPC-based approach for solving this is the measurement bandwidth which, for the SPC, is limited by the operating rate of DMD.
An approach to increasing the measurement rate is by using a multi-pixel architecture \cite{wang2015lisens, chen2015fpa,Mahalanobis:14}. One way to interpret such imagers is to think of each pixel on the sensor as an SPC. 
Hence, with the successful $128\times$ demonstrated in this paper, megapixel videos could potentially be achieved with the use of an $8 \times 8$ photodetector array. 
However, the very high-dimensionality of the recovered videos raises important computational challenges with regards to the use of optical flow-based  recovery algorithms.

%
%
%
%


%

\section*{Acknowledgments}
ACS was supported by the NSF grant CCF-­1117939.
LX, YL and KFK  were supported by ONR (N66001­11­1­4090), DARPA KeCoM (\#11­DARPA­1055) through Lockheed Martin, and Princeton MIRTHE (NSF EEC \#0540832).
RGB was supported by the grants NSF CCF-0431150, CCF-0728867, CCF-0926127,CCF-1117939, ARO MURI W911NF-09- 1-0383, W911NF-07-1-0185, DARPA N66001-11-1-4090, N66001-11-C-4092, N66001-08-1-2065, ONR N00014-12-1-0124 and AFOSR FA9550-09-1-0432.

\bibliographystyle{siam}

\end{document}